\documentclass[final]{cvpr}

\usepackage{graphicx}
\usepackage{amsmath}
\usepackage{amssymb}
\usepackage{booktabs}
\usepackage[export]{adjustbox}
\usepackage[breaklinks,colorlinks]{hyperref}

\usepackage[capitalize]{cleveref}
\crefname{section}{Sec.}{Secs.}
\Crefname{section}{Section}{Sections}
\Crefname{table}{Table}{Tables}
\crefname{table}{Tab.}{Tabs.}

\usepackage{multirow}
\usepackage[utf8]{inputenc}
\usepackage{times}

\usepackage[ruled]{algorithm2e}
\usepackage[noend]{algpseudocode}
\usepackage{enumerate}
\usepackage{mathtools}
\usepackage{caption}

\usepackage{array}
\newcommand{\PreserveBackslash}[1]{\let\temp=\\#1\let\\=\temp}
\newcolumntype{C}[1]{>{\PreserveBackslash\centering}p{#1}}
\newcolumntype{R}[1]{>{\PreserveBackslash\raggedleft}p{#1}}
\newcolumntype{L}[1]{>{\PreserveBackslash\raggedright}p{#1}}

\newcommand{\temp}[1]{\textcolor{gray}{#1}}
\newcommand{\LS}{\mathcal L}

\usepackage[capitalize]{cleveref}
\crefname{section}{Sec.}{Secs.}
\Crefname{section}{Section}{Sections}
\Crefname{table}{Table}{Tables}
\crefname{table}{Tab.}{Tabs.}
\newcommand\blfootnote[1]{%
  \begingroup
  \renewcommand\thefootnote{}\footnote{#1}%
  \addtocounter{footnote}{-1}%
  \endgroup
}

\newcommand{\comment}[1]{}

\newcommand{\R}{\mathbb{R}}

\newcommand{\Imm}{\operatorname{Imm}}

\newcommand{\Diff}{\operatorname{Diff}}
\def\argmin{{\operatorname{argmin}}}

\newcommand{\vol}{\operatorname{vol}}
\usepackage{tikz-cd}

\begin{document}
\title{BaRe-ESA: A Riemannian Framework for Unregistered Human Body Shapes}

\author{Emmanuel Hartman$^1$, Emery Pierson$^{2,4}$, Martin Bauer$^{1,3}$, Nicolas Charon$^3$, Mohamed Daoudi$^{4,5}$\\
Florida State University, Tallahassee, Florida, USA $^1$, \\ University of Vienna, Vienna, Austria$^2$\\ University of Houston, Houston, Texas, USA$^3$\\  Univ. Lille, CNRS, Centrale Lille, Institut Mines-Télécom, UMR 9189 CRIStAL, Lille, F-59000, France$^4$,\\ IMT Nord Europe, Institut Mines-Télécom, Univ. Lille, Centre for Digital Systems, Lille, F-59000, France$^5$}
\maketitle

\begin{abstract}
We present \textbf{Ba}sis \textbf{Re}stricted \textbf{E}lastic \textbf{S}hape \textbf{A}nalysis (BaRe-ESA), a novel Riemannian framework for human body scan representation, interpolation and extrapolation. BaRe-ESA operates directly on unregistered meshes, i.e., without the need to establish prior point to point correspondences or to assume a consistent mesh structure. Our method relies on a latent space representation, which is equipped with a Riemannian (non-Euclidean) metric associated to an invariant higher-order metric on the space of surfaces.  Experimental results on the FAUST and DFAUST datasets show that BaRe-ESA brings significant improvements with respect to previous solutions in terms of shape registration, interpolation and extrapolation. The efficiency and strength of our model is further demonstrated in applications such as motion transfer and random generation of body shape and pose.
\blfootnote{\textcolor{black}{This work was supported by ANR projects 16-IDEX-0004 (ULNE) and by ANR-19-CE23-0020 (Human4D); 
by NSF grants DMS-1912037, DMS-1953244, DMS-1945224 and DMS-1953267, and by FWF grant FWF-P 35813-N. Corresponding Author: M. Bauer (bauer@math.fsu.edu)}}
\end{abstract}
\section{Introduction}
Over the past decade there has been an increased interest in analyzing the morphological variability of the human anatomy. In particular, the issue of modeling and retrieving changes in full body shape and pose has a wide range of graphics applications spanning from 3D human modeling, to augmented and virtual reality for animated films and computer games. In these applications one is interested in a framework that allows for operations such as shape interpolation~\cite{eisenberger2021neuromorph}, -- the task of finding a plausible deformation between two given body scans -- shape extrapolation, -- the task of finding a plausible deformation given a body scan and a corresponding initial deformation (movement) -- and random shape generation. Furthermore, one is interested in obtaining a natural disentanglement of changes in the pose and in the shape of the human body~\cite{cosmo2020limp}, which in turn will allow for  operations such as motion transfer~\cite{Basset_3DV2021}.  

\noindent{\bf Motivation:}
A major challenge in the context of human body shape analysis is the registration problem, i.e. raw scans of human bodies are usually not equipped with any point correspondences and will, in general, not even admit a consistent mesh structure. Traditionally this issue has been tackled by finding new representations of the meshes with a consistent mesh structure and estimated point correspondences using an external framework such as functional maps~\cite{ovsjanikov2012functional, shrec19}. This data preprocessing step is then followed with an independent framework for the analysis of parametrized shapes, such as the As-Rigid-As-Possible (ARAP) energy~\cite{sorkine2007rigid} or the as isometric as possible energy~\cite{kilian2007geometric}. In recent years, in the context of general functional data analysis, it has been shown that such a separation into registration and subsequent analysis 
can introduce a significant bias into the resulting statistical analysis, and it has been acknowledged that a significant increase in performance can be obtained by using an unified approach~\cite{srivastava2016functional,jermyn2017elastic}. 

\begin{figure*}
\begin{tikzpicture}
	\draw[ultra thick]
    (-1,0) to (3,0);
    \draw[ultra thick]
    (0,4) to (4,4);
    \draw[ultra thick]
    (-1,0) to (0,4);
    \draw[ultra thick]
    (3,0) to (4,4);
	\coordinate[label=right:Latent space] (H) at (0,-.5);

	\draw[thick, ->] (-1, -.1) to (3, -.1);
	\coordinate[label=right: Shape] (shape) at (2.7, -.35);

	\draw[thick, ->] (-1.1, 0) to (-.1, 4);
	\coordinate[label=right: Pose] (shape) at (-1.1, 4);


    \draw[ultra thick]
    (5.5,0) to [out=15,in=165](9.5,0);
    \draw[ultra thick]
    (6.5,4) to [out=15,in=165](10.5,4);
    \draw[ultra thick]
    (5.5,0) to [out=100,in=225](6.5,4);
    \draw[ultra thick]
    (9.5,0) to [out=100,in=225](10.5,4);
    \coordinate[label={[align=left]Parameterized human shape space \\ with Sobolev Riemannian metric}] (H) at (7.,-.8);
    \node (image) at (5.8, 1.3) {
                    \includegraphics[height=0.1\linewidth]{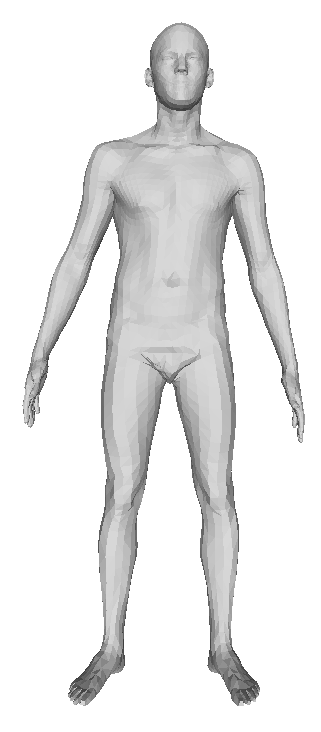}
            };
    \node[circle,fill,inner sep=2pt, orange] at (5.9,.5)   (template) {};    \draw[thick, dashed, gray, ->] (6.7, 7) to [out=-115, in=115] (5.5, 1.6);

    \node (image) at (9.4,1.3) {
                    \includegraphics[height=0.1\linewidth]{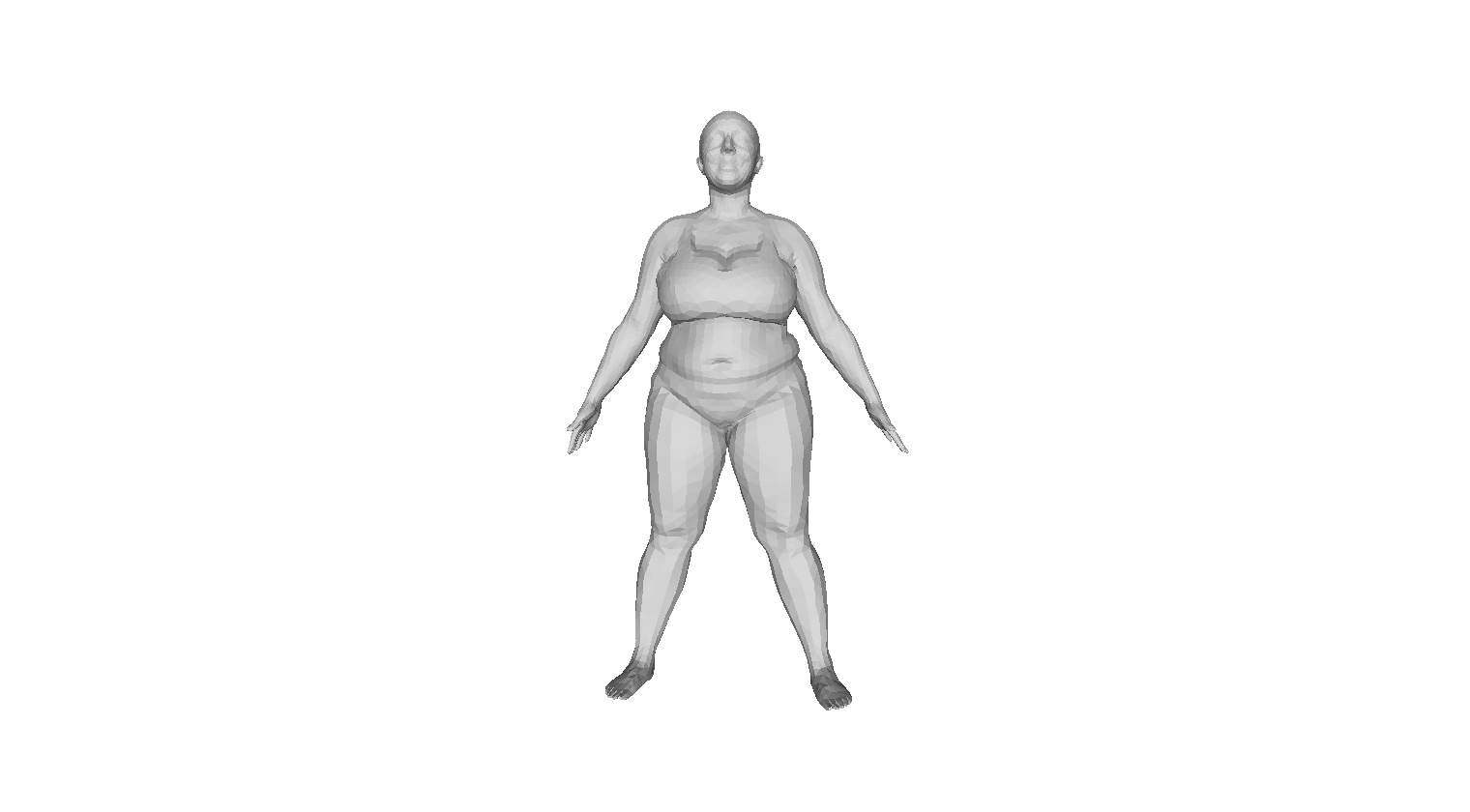}
            };
    \node[circle,fill,inner sep=2pt, orange] at (9.,.5)   (woman) {};
    \draw[thick, dashed, gray, ->] (7.7, 7) to [out=-115, in=135] (9.2, 1.6);

    \node (image) at (6.5,3.7) {
                    \includegraphics[height=0.1\linewidth]{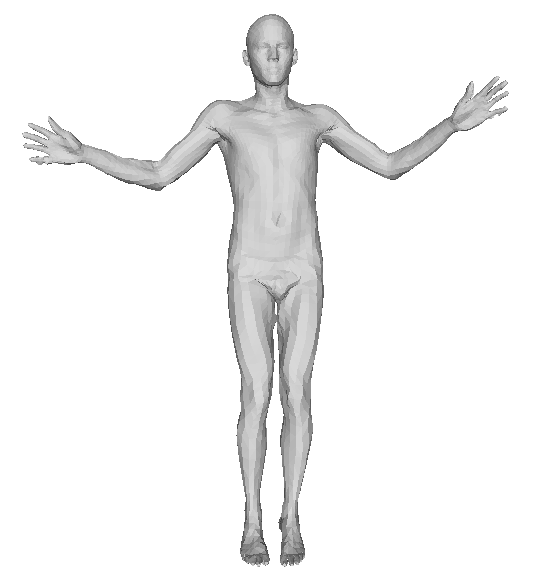}
            };
    \node[circle,fill,inner sep=2pt, orange] at (6.75,3.7)   (temp_up) {};
    \draw[thick, dashed, gray, ->] (6.7, 8) to [out=-115, in=135] (5.9, 3.6);

    \node (image) at (9.8,3.7) {
                    \includegraphics[height=0.1\linewidth]{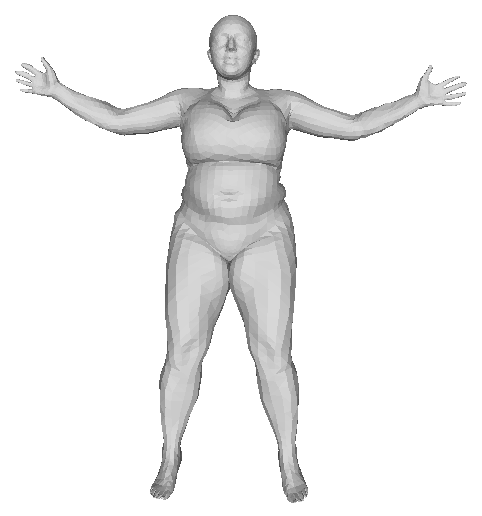}
            };
    \node[circle,fill,inner sep=2pt, orange] at (9.4,3.7)   (wom_up) {};
    \draw[thick, dashed, gray, ->] (8.3, 8) to [out=-15, in=55] (10., 4.3);
    
    \coordinate[label={[align=left]Human body latent code retrieval}] (H) at (8.,4.8);
    \node (image) at (10.,5.9) {
                    \includegraphics[height=0.1\linewidth]{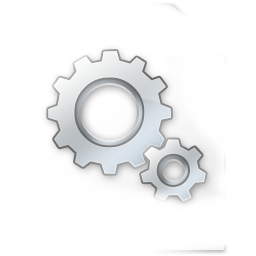}
            };

    \node[circle, fill, inner sep=2pt, cyan] at (-.3, .5) (lat_temp) {};
    \node[circle, fill, inner sep=2pt, cyan] at (.4, 3.5) (lat_temp_up) {};
    \node[circle, fill, inner sep=2pt, cyan] at (2.5, .5) (lat_temp_wom) {};
    \node[circle, fill, inner sep=2pt, cyan] at (3.3, 3.5) (lat_temp_wom_up) {};

    \draw[thick, red] (lat_temp) to [out=80,in=190] (lat_temp_wom_up);

    \draw[ultra thick, ->] (3.1,3.3) to [out=45, in=135] (6.5, 2.9);
    \coordinate[label={[align=left]Latent path}] (H) at (2.3,1.8);

    \coordinate[label={[align=left]Affine \\ decoder}] (H) at (4.6,2.8);

    \draw[ultra thick, ->] (5.7,1.3) to [out=-135, in=-45] (3., 0.9);
    \coordinate[label={[align=left]Pullback \\ metric}] (H) at (4.6,0.8);

    \node (image) at (4.3,7.7) {
                    \includegraphics[height=0.4\linewidth]{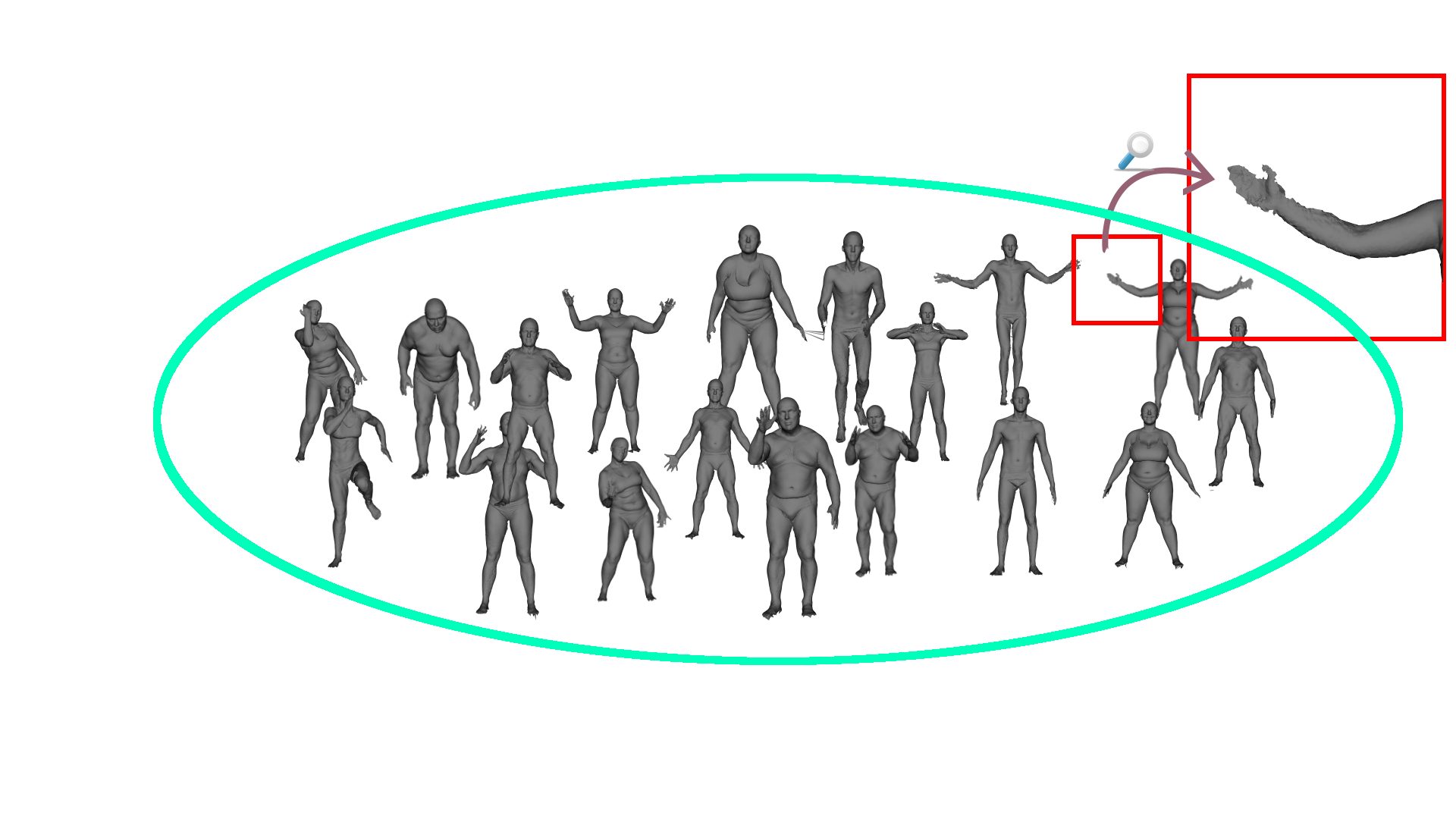}
            };

    \coordinate[label=right:Unparameterized human shapes] (unparam) at (3,10.2);
    \draw[thick, dashed] (10.7, 8.5) to (10.7, 0);

    \coordinate[label=right:Applications:] (app) at (11.,9);
    \node (image) at (13.,7.6) {
                    \includegraphics[height=0.08\linewidth]{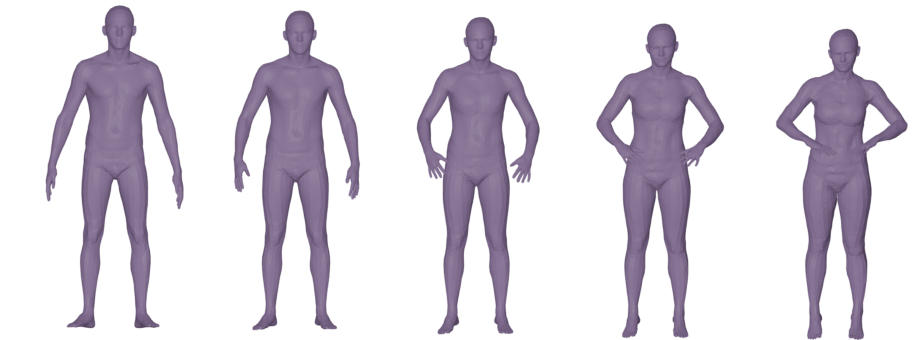}
            };

    \coordinate[label=right:Body scan Parameterization] (param) at (11., 6.5);
    \node (image) at (13.,5.5) {
                    \includegraphics[height=0.06\linewidth]{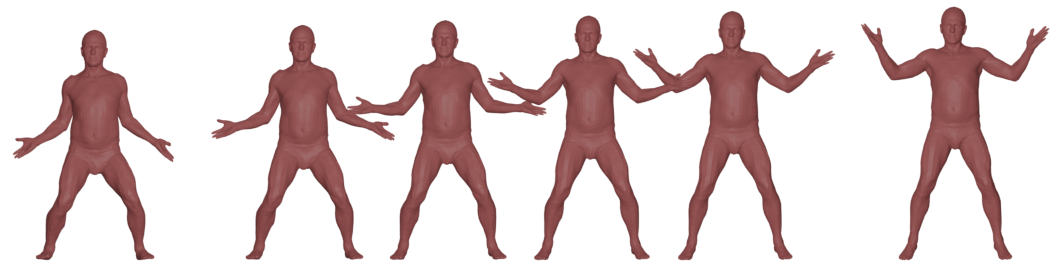}
            };

    \coordinate[label=right:Human shape interpolation] (interp) at (11., 4.5);
    \node (image) at (13.,3.5) {
                    \includegraphics[height=0.06\linewidth]{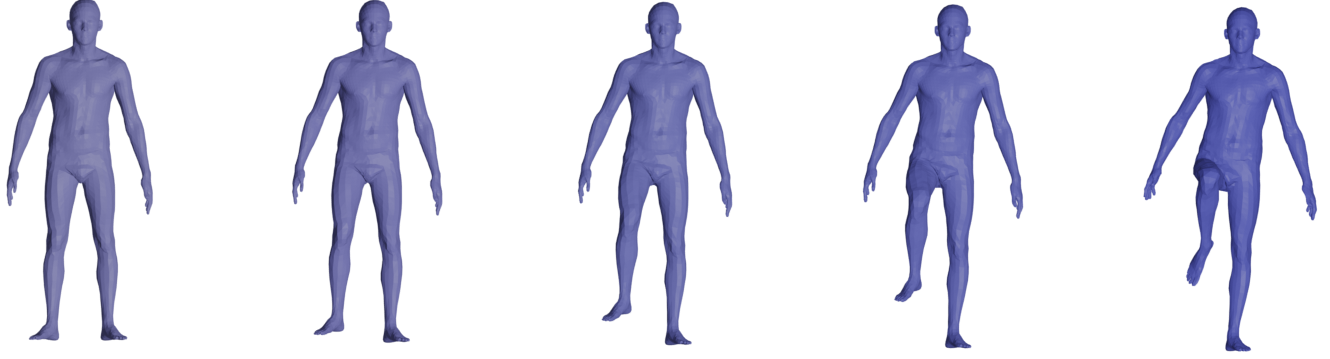}
            };
    \node (image) at (13.,2.5) {
                    \includegraphics[height=0.06\linewidth]{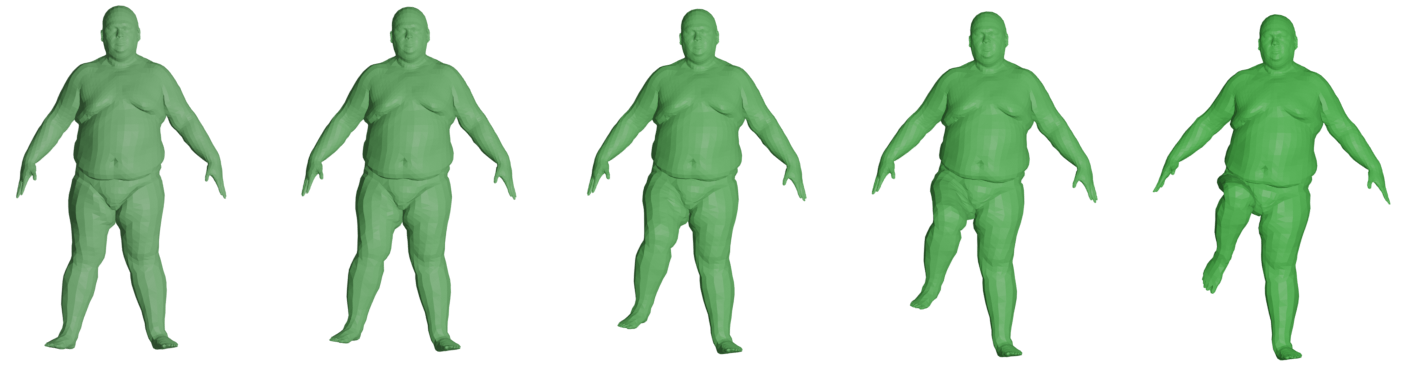}
            };
    \coordinate[label=right:Motion transfer] (trans) at (11., 1.5);
    \node (image) at (13.,0.6) {
                    \includegraphics[height=0.08\linewidth]{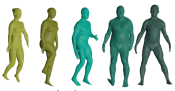}
            };
    \coordinate[label=right:Shape Generation] (gen) at (11., -0.5);

\end{tikzpicture}
\caption{Overwiew of our method. We seek to represent unparameterized human bodies, with different mesh connectivity, and possible noise or topological changes in a disentangled latent space. We define our latent space as the sum of pose and shape spaces. The paths in the latent space are not linear but curved, corresponding to geodesics in the paramaterized human body space. After retrieving the latent codes of the human bodies, we can use the space along with its Riemmanian metric to solve several problems in human body deformation: inter-extrapolation, motion transfer, shape generation.}\label{fig:pipeline}
\end{figure*}
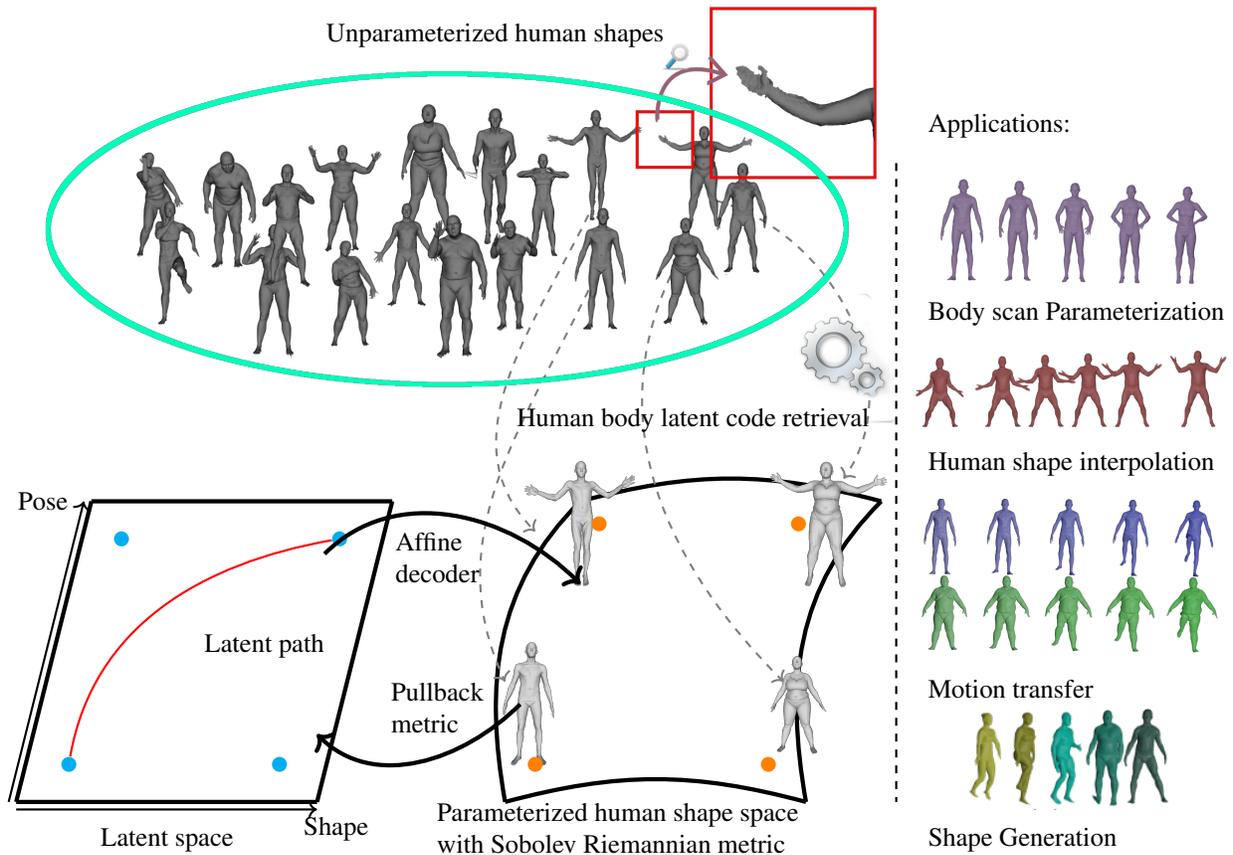

\noindent{\bf Contribution:}
Towards this aim we introduce a new pipeline that quantifies geometric differences between unregistered human body scans, i.e., that does not require prior point correspondences or consistent parametrization across the dataset. Furthermore, we are not only interested in a pure metric comparison of two individuals, but also in estimating plausible deformation processes from one human body to the other. To that end, we introduce a transformation model that allows to disentangle changes in the pose and in the shape of the human body so as to obtain realistic ways to interpolate from one scan to another, extrapolate a given motion or transpose it to a new individual and even to generate visually plausible random pose and shapes; see Figure \ref{fig:pipeline} for a schematic visualization of our framework. It is important to highlight that, through the use of the varifold representation and kernel metrics, our approach does not require having a consistent mesh structure across the dataset and performs well on human body scans with different numbers of vertices and even under the presence of topological noise (e.g. holes in the scans). Furthermore, unlike current deep learning frameworks for human body analysis, the encoding component in our approach relies on pre-trained deformation bases for the shape and pose changes but is coupled with a non-Euclidean metric in the latent space. Thus the training of our model is notably simple and does not require large amount of training data. Moreover, as our results suggest, it leads to better properties when it comes to the interpretability of human body paths and the generalization to unseen data. 


\section{Related Work}

\noindent{\bf Human body statistical spaces:}
Since Blanz~\etal's 3D Morphable Model for faces~\cite{blanz1999morphable}, statistical parametric models have been a preferred way for modeling 3D shape variations of human shapes (body, face and so on). In the particular domain of human bodies~\cite{anguelov2005scape, hasler2009statistical, pishchulin2017building}, the approaches consist of blending a skeleton transformation using several possible transformation models, along with a learned space of identity deformation. Recently, the SMPL model~\cite{SMPL:2015} is widely used for a variety of applications as it gives a valuable method of compressing the human body characteristics. The pipeline to retrieve the parameters of the SMPL model remains expensive and often requires manual intervention. As datasets of human body shapes grow larger, manual intervention becomes unfeasible and modern machine learning methods have been proposed to automate the processing of raw shape data.

\noindent{\bf Riemannian deformations of surfaces:} 
In order to deform human shapes a common approach is to optimize a deformation relative to a given energy. Multiple deformation energies have been used for defining the space of plausible deformations for human (or humanoid) shapes. Some project the points into Riemannian representation space~\cite{lipman2005linear, bansal2018lie}, or define energies from Riemannian metrics, such as the As-Rigid-As-Possible (ARAP)~\cite{sorkine2007rigid}, the as isometric as possible~\cite{kilian2007geometric} or the framework of Pierson~\etal~\cite{Pierson_2022_WACV}, who minimize a first order Riemannian energy with respect to deformations learned from training data of human motions. All these methods are, however, parameterization dependent and  thus point correspondences (and a consistent mesh structure) must be precomputed. 

The field of \textit{elastic shape analysis} (ESA)~\cite{srivastava2016functional}, which employs reparametrization invariant Riemannian metrics for defining deformation energies, has been proposed to overcome this drawback by solving for the registration (point correspondences) and deformation path in a unified fashion. This approach defines a shape as a point in the space of embeddings (or immersions) from a parameter space $M$ into $\R^3$, that can be quotiented by the action of reparameterizations of $M$ and the group of Euclidean transformations. The induced manifold is equipped with the inherited Riemannian metric from the space of parametrized surfaces and is then used to measure the distance between two given shapes as well as interpolation by computing a geodesic that connects them ~\cite{Kurtek_2012_PAMI,jermyn2017elastic, Su_2020_CVPRW, tumpach_gauge_2016,laga20224d}. These approaches, however, assume that the surfaces are given by analytic (spherical) representations. In the context of real data (raw body scans) this thus requires one to find such an analytic representation of the data in a pre-processing step~\cite{praun2003spherical,kurtek2013landmark}. In the presence of imaging noise in particular, this is a highly non-trivial task, that can introduce significant bias and error in subsequent statistical analyses. 

The varifold representation of surfaces~\cite{charon2013varifold, kaltenmark2017general}, can be used to define discrepancy loss functions between shapes that are not only invariant to parameterization, but also robust to scan inconsistencies such as potential noise or holes. They indeed have been apply successfully in the context of human shapes~\cite{pierson_eccv_2022}, although it was for sequence comparison and not deformation of surfaces. The approach of this paper follows some of the recent advances~\cite{Bauer_2021_IJCV, hartman2022elastic} that combine Riemannian elastic metrics with varifolds discrepancy terms and thereby overcome the aforementioned difficulties.

\noindent{\bf Data-driven human body latent spaces:}
The deep learning era has introduced multiple generative models such as the popular Generative Adversarial Networks (GAN)~\cite{NIPS2014_5ca3e9b1} and Variational AutoEncoders (VAE)~\cite{KingmaW13} that are able to model faithfully sample variations from a training set of images.
Geometric deep learning~\cite{bronstein2017geometric}, and in particular 3D autoencoders methods seek to extend those methods to 3D data. Those methods~\cite{bouritsas2019neural, lemeunier2022representation, huang2021arapreg, groueix2018b} proved to be successful in order to find a low dimensional representation of the space compared to classical algorithms such as PCA. In recent years, several  autoencoder frameworks have been proposed to represent human bodies in a low-dimensional latent vector space  while being independent of parameterization, thanks to different architecture such as PointNet~\cite{Qi_17_CVPR,gdvae_2019,cosmo2020limp,GLASSCvpr2022, groueix20183d} or more recently transformers, but their training cost remain high~\cite{TrappoliniNips2021}. By using several deformation energy losses in the training set such as geodesic distances~\cite{cosmo2020limp}, or ARAP~\cite{huang2021arapreg, GLASSCvpr2022}, these methods train non-linear encoder and decoder networks to map the low dimensional latent vector space to the space of human bodies and vice versa. This allows one to map linear paths in the latent space to paths of plausible human body motions. The main drawback of these methods comes from their reliance on large datasets of parameterized surfaces and their failure to sufficiently learn the non-linear map from the flat space to the space of human bodies thereby lacking in generalizability when confronted to unseen data. 

In contrast, our proposed pipeline, as illustrated by Figure \ref{fig:pipeline}, imposes an affine map, called the affine decoder, from a given low dimensional latent space to a corresponding space of human body transformations, which is based on the use of some pre-estimated bases to represent infinitesimal body shape and body pose deformations.  Recent approaches has demonstrated that regularizing latent space~\textcolor{black}{\cite{10.1007/978-3-031-20065-6_33, freifeld2012lie,atzmon2021augmenting}} can improve the computations of paths for 3D shapes. To do so, unlike aforementioned deep learning architectures, we do not rely on the standard Euclidean metric in the latent space but instead on the  pullback of a second order parametrization-invariant Sobolev metric in the space of deformation field. Thus, interpolating paths in latent space are no more straight lines, but are associated to (sub-Riemannian) geodesics in the surface space for this metric.

\label{sec:Related_Work}
\section{Riemannian Latent Space Method}\label{sec:model}
\noindent {\bf The space of human shapes:} In this article we model the space of all human shapes as a subset of the infinite dimensional space of template-based unparametrized surfaces, i.e., we view a human shape as an element of the quotient space $\mathcal S=\operatorname{Imm}/\operatorname{Diff},$ where $\operatorname{Imm}=\left\{ q\in C^{\infty}(\mathcal{T},\mathbb R^3): Tq \text{ is inj.}\right\}$
denotes the space of smooth maps (immersions)  and 
where $\operatorname{Diff}$
is the group  of all smooth, bijective maps (diffeomorphisms) on a human body template $\mathcal{T}$.

\noindent {\bf A Riemannian approach:} To define our framework for the analysis of human shapes we will resort to the field of Riemannian geometry, more precisely we will view the space 
$\mathcal S$ as an infinite dimensional manifold and equip it with a Riemannian metric. This will allow us to reduce tasks such as shape interpolation and extrapolation to geometric operations -- the geodesic initial and boundary value problem. 
In order to define a Riemannian metric on $\mathcal S$ we will follow the elastic shape analysis (ESA) paradigm of defining first a reparametrization \emph{invariant} Riemannian metric on the space of parametrized surfaces $\operatorname{Imm}$, that will then descend to a Riemannian metric on the quotient space~$\mathcal S$. Recall that a Riemannian metric $G$ is a family of inner products 
$$G_q:C^{\infty}(\mathcal{T},\mathbb R^3)\times C^{\infty}(\mathcal{T},\mathbb R^3)\to \mathbb R_+ $$
that depends smoothly on the foot point $q\in \Imm$. Here we have identified the tangent space $T_q\Imm$ (the space of deformation vectors) with the space of smooth vector fields over~$\mathcal{T}$.

\noindent {\bf Latent Space Model:} For applications it is natural to restrict this model to only consider a certain set of admissible deformations  applied to a template shape. In the context of shapes of human bodies, such a restriction ensures that the model only contains deformations that are "natural" i.e. deformations stay in the space of expected human body shapes. Therefore in our model, we restrict ourselves to surfaces $q$ that can be written as linear combinations of a basis of admissible body pose deformations (shape basis), $\{h_i\}_{i=1}^{m}$, and a basis of admissible body type deformations (motion basis), $\{k_i\}_{i=1}^n$. Thus,  all of the shapes we consider are determined by the latent space $\LS\subset \mathbb R^{n+m}$ of coefficients for the combined basis. This latent space model is thus related to the space of immersions by the mapping $F:\LS\to \Imm$
via 
$$F(\alpha^j)=\bar q+\sum_{i=1}^m \alpha^j_i h_i+\sum_{i=m+1}^{m+n} \alpha^j_i k_{i-m}.$$
where $\bar q$ is our parameterized template human body shape, $h_i$ is a basis of (realistic) body type (identity) deformations and where $k_i$ is a basis of (realistic) body pose  deformations (movements). This choice of bases is a crucial ingredient to obtain a good performance of the proposed method.  In the present work we will construct them in a data driven way, which will be described in more details in Section~\ref{sec:basis_construction}. We then equip our latent space with a non-euclidean metric by defining the pullback of $G$ via $F$. In particular,  for $\alpha \in \LS$ and $\beta,\eta\in\LS$, we define the metric on the latent space by
\[\overline{G}_{\alpha}(\beta,\eta):= G_{F(\alpha)}(F(\beta )-\overline{q},F(\eta )-\overline{q}).\]

\noindent {\bf Choice of Metric:} There is a variety of different Riemannian metrics $G$ satisfying the required invariance properties that have been proposed in the literature. In the current article we choose a metric from the family of second order Sobolev metrics with six parameters proposed in \cite{hartman2022elastic}. We will not present the exact formula of the Riemannian metric here, but refer the interested reader to supplementary material. For the purpose of the present article we only mention the following fundamental properties of our choice of Riemannian metric,  which are central for the proposed applications: 1) the second order term in the metric provides enough regularity to prevent the occurence of singularities along the solution of the interpolation and extrapolation problem described above, 2) the parameters in the metric provide enough flexibility to enforce certain behaviors of the resulting optimal deformations, eg. as isometric as possible deformations and 3) they naturally can be extended to the space of triangulated meshes using the methods of discrete differential geometry.

\noindent {\bf Interpolation as a geodesic boundary value problem:}
The shape interpolation problem between two surfaces (human shapes) is the task of finding an optimal deformation (path of immersions) between the two given surfaces. In our Riemannian setup this reduces to minimizing the energy
$$E(\alpha)=\int_0^1 \overline{G}_{\alpha}(\partial_t\alpha,\partial_t\alpha) dt$$
over all paths $\alpha:[0,1]\to \LS$ and diffeomorphisms $\varphi\in \Diff$ such that $F(\alpha)(0)=q_0$ and $F(\alpha)(1)=q_1\circ \varphi$, where $q_0$ and $q_1$ are arbitrarily chosen parametrizations of the unparametrized shapes. 
In order to apply this  model to real data the main difficulty is the action of the diffeomorphism group on the endpoint constraint and the fact that raw body scans are, in general, not equipped with a consistent mesh structure, i.e., different scans can have different resolution, different mesh structures or even involve imaging errors resulting in e.g. holes or missing parts.

To circumvent these difficulties we 
will instead consider a relaxed formulation of the BVP given by the energy 
\begin{multline}
   \tilde E(\alpha)=\int_0^1 \overline{G}_{\alpha}(\partial_t\alpha,\partial_t\alpha) dt\\+\lambda \Gamma(F(\alpha)(0),q_0)+\lambda \Gamma(F(\alpha)(1),q_1), \label{eq:sym_match_energy}
\end{multline}
where $\Gamma$ is a reparametrization blind similarity measure, i.e., $\Gamma$ satisfies the fundamental property $\Gamma(q_0,q_1)=0$ if and only if $q_0$ and $q_1$ represent the same shape, i.e., there exists a $\varphi \in \Diff$ such that $q_0=q_1\circ\varphi$. We have thus reduced the geodesic boundary value problem to an unconstrained minimization problem over all paths  $\alpha:[0,1]\to \LS$. There are various different possibilities for this similarity term, such as the Hausdorff or Chamfer distance. We will  rely on a similarity term derived from geometric measure theory, namely kernel metrics on the space of varifolds previously used in e.g. \cite{kaltenmark2017general}. An important advantage of this framework is the fact that the resulting fidelity loss function remains differentiable with respect to the positions of the shapes' vertices. Additionally, after discretization, the varifold distance does not require the given meshes to have a consistent mesh structure which allows us to compare human body scans with different numbers of vertices and even different topologies, e.g. in the case of holes in the scans.

We then discretize the path $\alpha$ in time and compute the path energy using finite differences. Further, we may discretize the varifold data similarity term as in \cite{kaltenmark2017general}. Details of such a discretization are included in the supplementary materials. Thus, we have reduced the infinite dimensional minimization problem to a finite dimensional one, with the variables being the (discrete) path of latent variables. We tackle the resulting unconstrained non-convex minimization problem using the L-BFGS algorithm. 

\noindent {\bf Extrapolation as a geodesic initial value problem:}
The shape extrapolation problem consists in predicting the future evolution of a surface (human body) given an initial deformation direction. In our Riemannian framework this reduces to solving the geodesic equation  with given initial condition $q(0)=q_0$ (the initial pose) and $\partial_t q(0)=h$ (the direction of deformation). The geodesic equation is the first order optimality condition of the energy functional; it is non-linear PDE, that is second order in time $t$ and forth order in space. For the exact formula of this equation, which is rather lengthy and not particularly insightful, we refer the interested reader to the literature, see eg.~\cite{bauer2011sobolev}. To solve such initial value problems in our latent space, we modify methods of discrete geodesic calculus~\cite{rumpf2013discrete} for our setting. We approximate the geodesic starting at $\alpha^0$ in the direction of $\beta$ with a PL path with $N+1$ evenly spaced breakpoints. At the first step, we set $\alpha^1=\alpha^0+\frac{1}{N}\beta$ and find $\alpha^2$ such that $F(\alpha^1)$ is the geodesic midpoint of  $F(\alpha^0)$ and  $F(\alpha^2)$, i.e., we solve for $\alpha^2$ such that 
\begin{equation*}
\alpha^1 = \underset{\Tilde{\alpha}}{\argmin} [\overline{G}_{\alpha^0}(\beta_0,\beta_0)+\overline{G}_{\Tilde{\alpha}}(\Tilde{\beta},\Tilde{\beta})]
\end{equation*} where $\beta_0=\Tilde{\alpha}-\alpha^0$ and $\Tilde{\beta}=\alpha^2-\Tilde{\alpha}$.
Differentiating with respect to $\Tilde{\alpha}$ and evaluating the resulting expression at $\alpha^1$, we obtain the system of equations 
\begin{multline}
    2\overline{G}_{\alpha^0}(\beta_0,h_i)-2\overline{G}_{\alpha^1}(\Tilde{\beta},h_i)+ D_{\alpha^1}\overline{G}_{\cdot}(\Tilde{\beta},\Tilde{\beta})_i=0, \\
    2\overline{G}_{\alpha^0}(\beta_0,k_i)-2\overline{G}_{\alpha^1}(\Tilde{\beta},k_i)+ D_{\alpha^1}\overline{G}_{\cdot}(\Tilde{\beta},\Tilde{\beta})_{i+m}=0  \label{eq:first_order_condition_ivp}
\end{multline}
where $\{h_i,k_i\}$ is our basis of deformations as introduced above. We denote the system of equations in~\eqref{eq:first_order_condition_ivp} by $\Phi(\alpha^2; \alpha^1, \alpha^0) = 0$, where we stress again that $\alpha^0$ and $\alpha^1$ are here fixed and known. We solve this system of equations for $\alpha^2$ using a nonlinear least squares approach,
\begin{equation*}
    \alpha^2 = \underset{\Tilde{\alpha}}{\argmin} \| \Phi(\Tilde{\alpha}; \alpha^1, \alpha^0) \|_2^2.
\end{equation*}
 We repeat this process $N-1$ times, thereby constructing the discrete solution up to time $t=1$.

\subsection{A data-driven basis of deformations}\label{sec:basis_construction}
To construct the bases of movements and body type deformations we interpret registered mesh sequences of motions as paths in shape space whose tangent vectors are implicitly restricted to the space of valid motions of a human body. The tangent vectors of those sequences are used as training data, on which we perform principle component analysis to obtain a tractable yet expressive basis for the valid pose deformations of a human body. We then collect meshes of the same pose from each identity (generally available in T or A-pose) and compute the (unrestricted) pairwise geodesics between these meshes with respect to our second-order Sobolev metric, where we use the Pytorch implementation of~\cite{hartman2022elastic}.
Note that these meshes show only moderate deformations and thus there are no difficulties with applying the unrestricted matching algorithm. We then collect the tangent vectors to these paths and perform again PCA to define our basis of shape deformations. \textcolor{black}{For the analysis of  pre-registered human body motions a similar procedure has been used in~\cite{Pierson_2022_WACV}}. \textcolor{black}{We discuss the limitations and benefits of this process in Section 8 of the supplementary material.}

\noindent{\bf Parameter selection:} 
The resulting basis size for our model is as follows (the number of elements was chosen experimentally): the motion basis has $n=130$ elements, whereas the basis for the body type variation has only  $m=40$ elements. The coefficients for the $H^2$-metric were chose to enforce close to isometric deformations that allow for some stretching and shearing to allow change in body type. Further, we added a small coefficient to the second-order term to further regularize the deformations. The final six parameters for the $H^2$-metric are set to $(1, 1000, 100, 1, 1, 1)$ 
We perform sequential minimizations where the parameter $\sigma$ of the varifold term is decreased from $.4$ to $.025$ and the balancing term $\lambda$ is increased from $10^{2}$ to $10^{8}$. A pseudo code of our main algorithm is available in the supplementary material.

\section{Results}
In this section we demonstrate the accuracy of our framework in five different experiments:
the registration of human body scans, the interpolation and extrapolation of human body motions, random 
generation and motion transfer. 
\subsection{Datasets}
\noindent{\bf Training dataset:}
To construct our basis we will make use of the publicly available 
Dynamic FAUST (D-FAUST)~\cite{dfaust:CVPR:2017} dataset. 
This dataset contains high quality scans, along with registered meshes that will be used as training data. More specifically
D-FAUST~\cite{dfaust:CVPR:2017} is a 4D scan dataset of human motions. 10 individuals performed at most 14 in-place motions, recorded using multi view cameras and high quality 4D scanner, sampled at 60 Hz. Due to the high speed of the recording, D-FAUST scans contain several singularities in the reconstructed surface, such as holes or even artificial objects (eg. parts of walls). The SMPL mesh is registered using a registration pipeline based on image texture information, and a novel body motion model, allowing to have the registrations corresponding to each scan.
A set of 7 long range sequence, divided in 10 representative mini-sequences, are left for testing (see supp. material for the video of these sequences). 

\noindent{\bf Test dataset:}
In addition to these 7 sequences from D-FAUST we tested our algorithm on the static FAUST dataset~\cite{Bogo:CVPR:2014}. This is a 3D static scan dataset designed for human mesh registration, that contains scans of minimally clothed humans, similar to the ones in the D-FAUST dataset. Ten individuals (different from the D-FAUST training set) performed 30 different poses recorded using a high quality 3D scanner. We selected 9 unregistered poses of the training set that show no rotations, and use them as a testing set. We chose this dataset rather than other possible ones~\cite{anguelov2005scape, shrec19} because the geometric closeness of minimally clothed humans allows us to evaluate the geometric quality of registrations (as opposed to the case of clothed humans) and it allows to test robustness against topological noise.

\subsection{Evaluation and comparison}\label{sec:eval}

Before presenting our results, we discuss our procedure for evaluating and comparing different methods. We perform a thorough comparison of our approach with other state-of-the-art approaches for human body analysis that rely on latent space learning for  registration, interpolation, and extrapolation tasks. To keep the section more focused, we deliberately restrict to those, and do not consider other methods that can potentially tackle the same tasks but without a low dimensional latent space~\cite{eisenberger2021neuromorph}, or that are specifically designed for other tasks~\cite{GLASSCvpr2022}. We compare our approach to LIMP~\cite{cosmo2020limp}, which models shape deformations using a variational auto-encoder with geodesic constraints; ARAPReg~\cite{huang2021arapreg}, which models deformations using an auto-decoder with regularization through the as rigid as possible energy; and 3D-Coded~\cite{groueix2018b}, which is similar to LIMP but with lighter training and without geometric loss regularization. LIMP and 3D-Coded both utilize a PointNet architecture as an encoder, which enables invariance to parameterization. On the other hand, ARAPReg recovers latent vectors within a registered setting utilizing the $L^2$ metric, which assumes that the target meshes possess identical mesh structure as the model's output. To make this framework viable for our application we replace the $L^2$-metric by the varifold distance thereby extending ARAPReg to unregistered point clouds. We trained all three networks on the D-FAUST dataset using reported training details from the respective papers. We used 3D-Coded only for comparison in the latent code retrieval experiments as its interpolation and extrapolation results were notably bad. 
The training of those methods for the experiments presented in this paper were carried out using the Grid'5000 testbed, supported by a scientific interest group hosted by Inria and including CNRS, RENATER and several Universities as well as other organizations (see \url{https://www.grid5000.fr}).

In our experiments we evaluate the quality of the results using different similarity measures (distances) between the outputs of the different methods and the original scan. The ``shape'' matching is evaluated by comparing each method against the original scans using three different remeshing invariant similarity measures.  First, we evaluate the methods using the varifold metric introduced above. As our method minimizes this distance during the registration process, we include two additional metrics to avoid bias: the widely used  Hausdorff distance, which provides a good insight for the quality of a mesh reconstruction, but can be sensitive to single outliers present in low quality scans and the Chamfer distance~\cite{fan2017point, groueix20183d}, which is more robust to such outliers.  

In our first experiment -- latent code retrieval -- we will in addition evaluate the quality of the obtained point correspondences -- in this section we use data with given ground truth point correspondences. Therefore we will compute the mean squared error of the each method to the ground truth registrations of the testing set. Unfortunately, one method (LIMP) does not return the same mesh structure as the ground truth registrations and thus we could not compare it this way. For a detailed description of all these evaluation metrics we refer to the supplementary material.

\subsection{Mesh Invariant Latent Code Retrieval}\label{sec:latent_code_retrieval}
Given a scan, $q_1$ with arbitrary mesh structure, we retrieve the latent code 
of the shape class of $q_1$ by performing a relaxed geodesic matching problem between $\overline{q}$ and $q_1$ with $T$ time steps. This produces a geodesic path from the template to the shape class of the target mesh, thus the endpoint $\alpha^T$ is the latent code 
of the shape class of $q_1$.

\begin{table}[]
    \centering
    \small
    \begin{tabular}{c}
        \begin{adjustbox}{max width=\textwidth}
        \begin{tabular}{l||c||c|c|c}
        &MSE&Hausdorff & Chamfer & Varifold    \\\hline
        LIMP&NA&0.23 & 0.098 & 0.073  \\
        ARAPReg& 0.035&0.11 & 0.117 & 0.021 \\
        3D-Coded& 0.053 &\textbf{0.07} & 0.020 & 0.023 \\
        BaRe-ESA& \textbf{0.015}&0.08& \bf{0.019} & {\bf 0.014} 
        \end{tabular}
        \end{adjustbox}
    \end{tabular}
    \caption{Shape registration and reconstruction results. Each method is trained on D-FAUST and tested on Faust. Where applicable, we compute the mean squared error (MSE) between each method's outputs and the ground truth registration of FAUST. We compute the reconstruction errors between the outputs of the methods and the original scans.}
    \label{tab:faust}
\end{table}

\begin{figure}
    \centering
    \small
    \begin{tabular}{cm{6cm}}
        \begin{tabular}{l}BaRe-ESA \end{tabular}&\includegraphics[width=\linewidth]{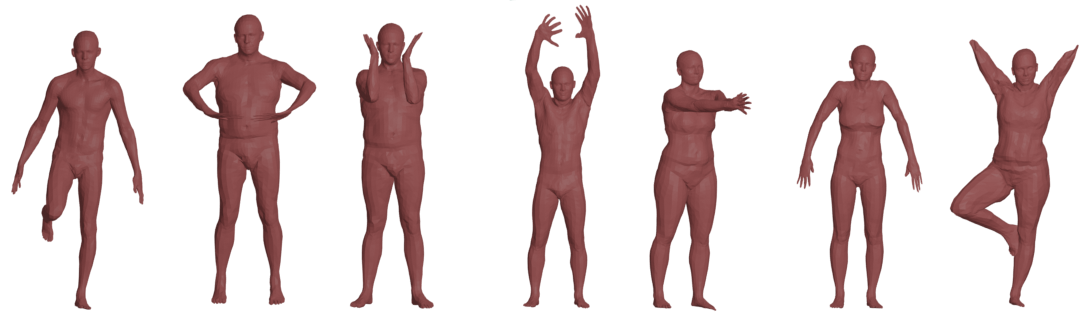}  \\\hline
        \begin{tabular}{l}LIMP \end{tabular}&\includegraphics[width=\linewidth]{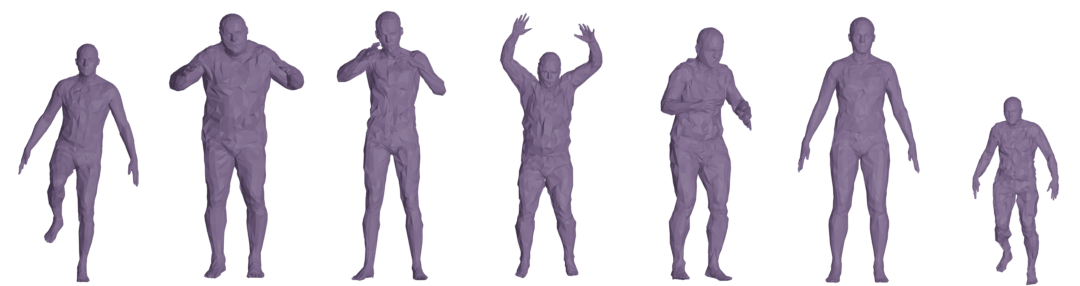}  \\\hline
        \begin{tabular}{l}ARAPReg \end{tabular}&\includegraphics[width=\linewidth]{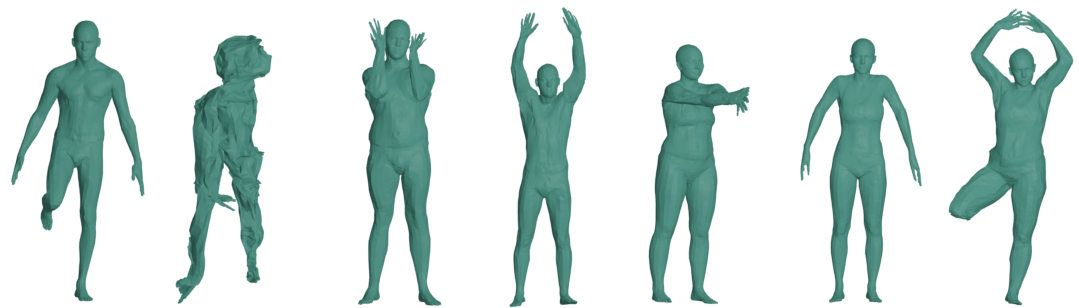} \\\hline
        \begin{tabular}{l}3D-Coded \end{tabular}&\includegraphics[width=\linewidth]{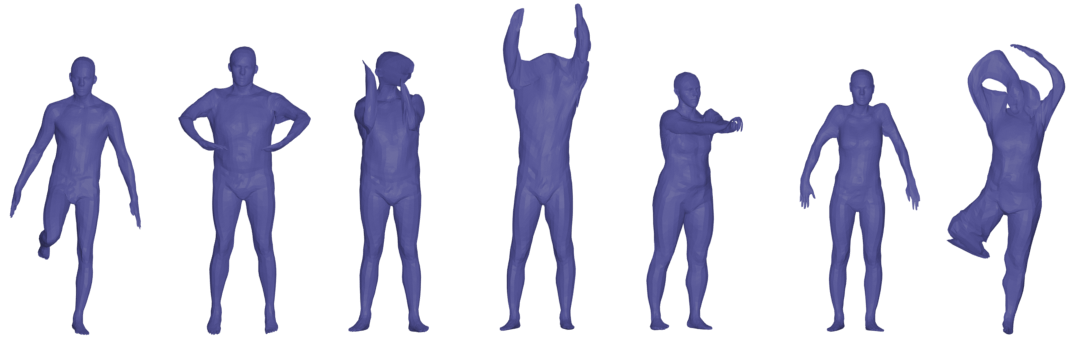} \\\hline
        \begin{tabular}{l}GROUND\\TRUTH \end{tabular}& \includegraphics[width=\linewidth]{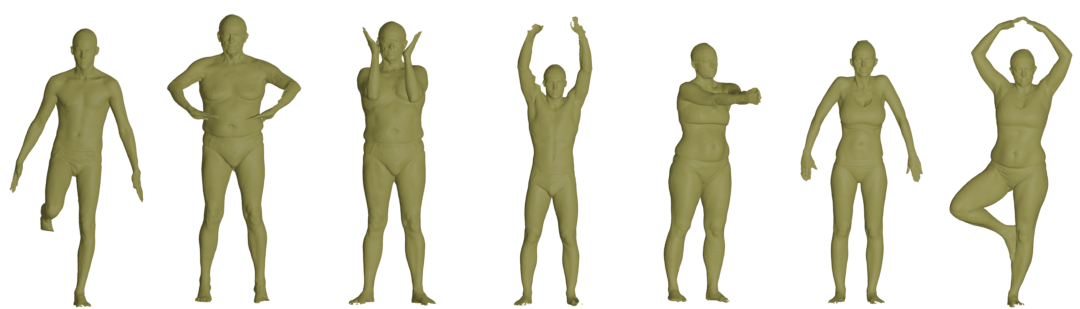}  \\
    \end{tabular}
    \caption{Registration of five elements of FAUST using three methods trained on D-FAUST. Both LIMP and ARAPreg fail to generalize to the new data and do not represent the ground truth in certain examples.  BaRe-ESA consistently produces a decent representation in all examples, with a failure case displayed in the last column.}
    \label{fig:registration}
\end{figure}

To demonstrate the effectiveness of our mesh latent code retrieval algorithm we test on the 90 meshes from the unregistered FAUST dataset (recall that this dataset was not contained in the training process). In this experiment, we construct latent code representations with BaRe-ESA, LIMP, 3D-Coded, and ARAPreg and measure the distance from the reconstructed meshes to the original scans using the evaluation methods outlined in Section~\ref{sec:eval}. In ~\Cref{fig:registration} we present a qualitative comparison of the obtained results.  A quantitative comparison of the performance of the different methods is presented in~\Cref{tab:faust} where we demonstrate that BaRe-ESA significantly outperforms the mesh autoencoder methods with respect to the registration and reconstruction evaluation metrics. In the supplementary material we also discuss the computational cost of our method.

\begin{table*}[t]
        \centering
        \scriptsize
        \setlength\tabcolsep{2pt} 
        \begin{tabular}{c}
            \begin{adjustbox}{max width=\textwidth}
                \aboverulesep=0ex
                \belowrulesep=0ex
                \renewcommand{\arraystretch}{1.0}
                \begin{tabular}[t]{c|ccc|ccc|ccc|ccc|ccc|ccc}
                \multicolumn{1}{c}{} & 
                \multicolumn{9}{c}{Interpolation} & 
                \multicolumn{9}{|c}{Extrapolation} \\
                \cmidrule{2-19}
                \multicolumn{1}{c}{} &
                \multicolumn{3}{c}{Hausdorff} &
                \multicolumn{3}{c}{Chamfer} &
                \multicolumn{3}{c}{Varifold} &
                \multicolumn{3}{c}{Hausdorff} &
                \multicolumn{3}{c}{Chamfer}  &
                \multicolumn{3}{c}{Varifold} \\
                & LIMP & ARAPReg & BaRe-ESA  & LIMP & ARAPReg & BaRe-ESA & LIMP & ARAPReg & BaRe-ESA & LIMP & ARAPReg & BaRe-ESA& LIMP & ARAPReg & BaRe-ESA& LIMP & ARAPReg & BaRe-ESA \\
                    \midrule
                    punching &
                     0.13 & 0.12 & \textbf{0.081} & 
                     0.029 & \textbf{0.020} &0.04&
                     0.060 & 0.034 & \textbf{0.022} &
                     0.140 & 0.368 &  \textbf{0.102} &
                     \textbf{0.030} & 0.112 &0.085&
                     0.066 & 0.097 & \textbf{0.025}  \\
                     
                     running on spot &
                     0.28 & 0.14 & \textbf{0.076} & 
                     0.112 & 0.080 & \textbf{0.069}&
                     0.072 & 0.052 & \textbf{0.025} &  
                     0.287 & 0.309 & \textbf{0.152} &
                     0.11 & 0.177 & \textbf{0.125} &
                     0.071 & 0.079 & \textbf{0.027} \\
                    
                    running on spot b &
                     0.20 & 0.13 & \textbf{0.082} &  
                     0.125 & 0.101 & \textbf{0.045} &
                     0.068 & 0.040 & \textbf{0.025}  &
                     0.264 & 0.415 & \textbf{0.222} &
                     0.138 & 0.179 & \textbf{0.116} &
                     0.071 & 0.083 & \textbf{0.027} \\
                     
                     shake arms &
                     0.34 & 0.33 & \textbf{0.078} & 
                     0.063 & \textbf{0.049} &0.076&
                     0.061 & 0.031 & \textbf{0.025}  &
                     0.410 & 0.832 & \textbf{0.273} &
                     \textbf{0.080} & 0.283 &0.171 &
                     0.066 & 0.083 & \textbf{0.027} \\
                     
                     chicken wings &
                     0.18 & 0.15 & \textbf{0.093} &
                     0.101 & 0.083 & \textbf{0.04}&
                     0.062 & 0.029 &  \textbf{0.016} &  
                     0.182 & 0.395 & \textbf{0.092} &
                     0.11 & 0.189 & \textbf{0.092}&
                     0.072 & 0.081 & \textbf{0.018} \\
                     
                     knees &
                     0.060 & 0.35 & \textbf{0.051} &
                     0.182 & 0.266 & \textbf{0.036} &
                     0.097 & 0.066 & \textbf{0.016}  &
                     1.13 & 0.282 & \textbf{0.104} &
                     0.23& 0.205 & \textbf{0.081} &
                     0.321 & 0.072 & \textbf{0.016} \\
                     
                     knees b &
                     0.52 & \textbf{0.054} & 0.084 &
                     0.168 & 0.107 & \textbf{0.029} &
                     0.067 & 0.019 &  \textbf{0.016}  &
                     0.489 & 0.319 & \textbf{0.244} &
                     0.17 & 0.215 & \textbf{0.065} &
                     0.066 & 0.071 & \textbf{0.017} \\
                     jumping jacks &
                     0.12 & 0.32 & \textbf{0.046} & 
                     0.074 & 0.054 & \textbf{0.014} &
                     0.070 & 0.044 & \textbf{0.015}  & 
                     0.195 & 0.645 & \textbf{0.091} &
                     0.076 & 0.302 & \textbf{0.072} &
                     0.068 & 0.086 & \textbf{0.027}\\
                     jumping jacks b &
                     0.40 & 0.11 & \textbf{0.062} & 
                     0.111 & 0.086 & \textbf{0.009}&
                     0.076 & 0.030 & \textbf{0.025}  &
                     0.492 & 0.570 & \textbf{0.122} &
                     0.12 & 0.324 & \textbf{0.061} &
                     0.10 & 0.083 & \textbf{0.029}\\
                     one leg jump &
                     0.16 & \textbf{0.062} & 0.07 &
                     0.138 & 0.109 & \textbf{0.067}&
                     0.068 & 0.025 & \textbf{0.024} &
                     0.175 & 0.363 & \textbf{0.167} &
                     0.143 & 0.249 & \textbf{0.136} &
                     0.069 & 0.078 & \textbf{0.025}\\
                     \midrule
                     mean &
                     0.30 & 0.19 & \textbf{0.072}  &
                     0.106 &0.097 & \textbf{0.042} &
                     0.070 & 0.039 & \textbf{0.020} & 
                     0.391 & 0.452 & \textbf{0.157} &
                     0.118 & 0.214 & \textbf{0.100}&
                     0.103 & 0.082 & \textbf{0.023} 
            \end{tabular} 
            \end{adjustbox}
          
        \end{tabular}
        \vspace{-3pt}
        \caption{Full interpolation and extrapolation comparison on 10 D-FAUST sequences. The Hausdorff, Chamfer and varifold distance are computed against ground truth sequences.\vspace{10pt}}
        \label{tab:interp_extrap}
        \vspace{-18pt}
    \end{table*}

\subsection{Interpolation} 
\begin{figure}
    \centering
    \small
    \begin{tabular}{cm{5.75cm}}
        \begin{tabular}{l}BaRe-ESA \end{tabular}&\includegraphics[width=\linewidth]{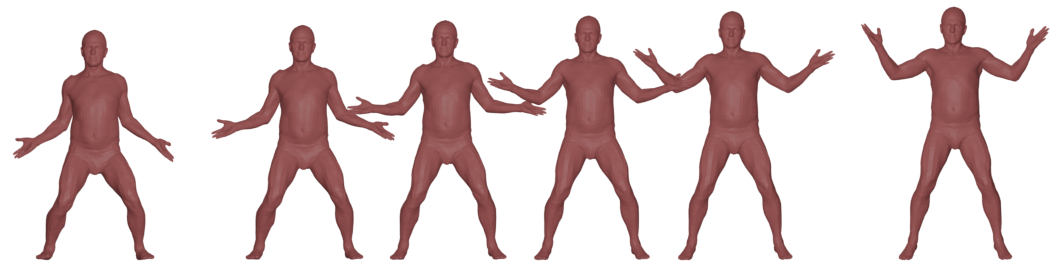}  \\\hline
        \begin{tabular}{l}LIMP \end{tabular}&\includegraphics[width=\linewidth]{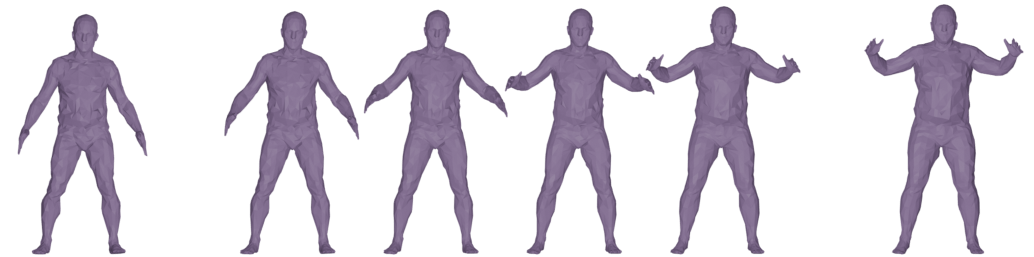}  \\\hline
        \begin{tabular}{l}ARAPreg \end{tabular}&\includegraphics[width=\linewidth]{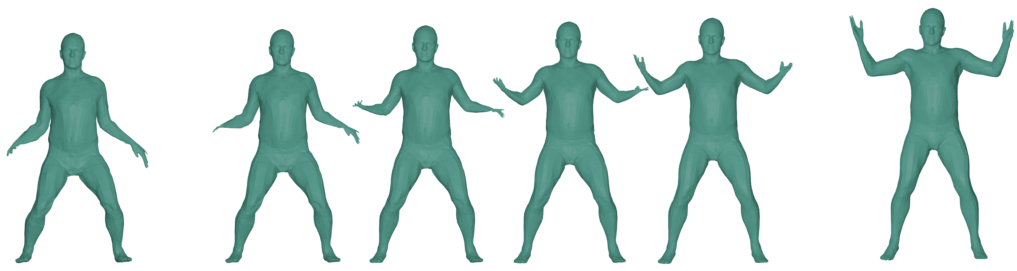} \\\hline
        \begin{tabular}{l}GROUND\\TRUTH\end{tabular}& \includegraphics[width=\linewidth]{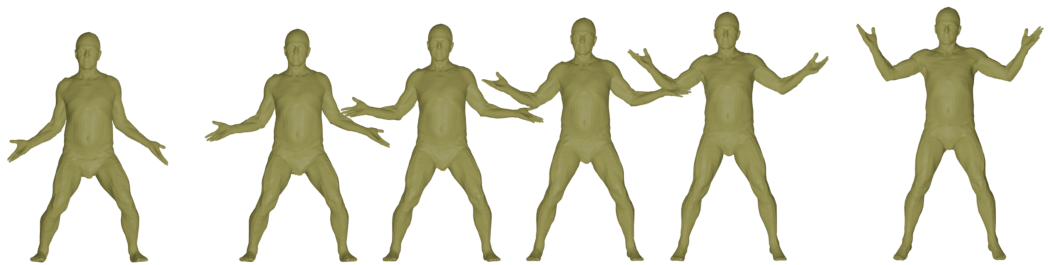}  \\
        &$\,\,\underbracket{\hspace*{.75cm}}_{\text{Source}}\,\quad\underbracket{\hspace*{3cm}}_{\text{Interpolation}}\,\quad\underbracket{\hspace*{.75cm}}_{\text{Target}}$
    \end{tabular}
    \caption{Interpolation results comparison between our method, LIMP, ARAPReg and the Ground Truth from D-FAUST. While the path produced by LIMP does not properly register the endpoints and the path produced by ARAPreg does not stay in the space of human bodies, BaRe-ESA successfully produces a path of human shapes whose endpoints match the source and target shapes.  }
      \label{fig:interpolation}
\end{figure}

Next we turn our attention to the interpolation problem, i.e., the task of constructing a deformation between two different human body poses, that follows a ``realistic'' motion pattern.
In our Riemannian setup, as described previously, this corresponds to solving the geodesic boundary value problem, i.e., we need to minimize the discrete energy given in~$\eqref{eq:sym_match_energy}$. We use the start and end point of our 10 test mini-sequences as the input for our interpolation problem. This allows us to compare the obtained results to the full mini-sequences, seen as a ground truth motion (see the supplementary material for their corresponding animations). In ~\Cref{fig:interpolation}, we compare the results of our method, ARAPReg, and LIMP with the ground truth for one mini-sequence. Our method is successful at recovering the latent codes that represent the endpoints and producing interpolations that remain in the space of human shapes. We further perform a quantitative comparison of the methods by measuring the distance to the ground-truth sequences at each break point with respect to the evaluation metrics given in Section \ref{sec:eval}; these results are displayed in~\Cref{tab:interp_extrap}. One can clearly observe that our method again outperforms the other methods both qualitatively and quantitatively.  

\subsection{Extrapolation}
\begin{figure}
    \small
    \begin{tabular}{cm{5.75cm}}
        \begin{tabular}{l}BaRe-ESA \end{tabular}&\includegraphics[width=\linewidth]{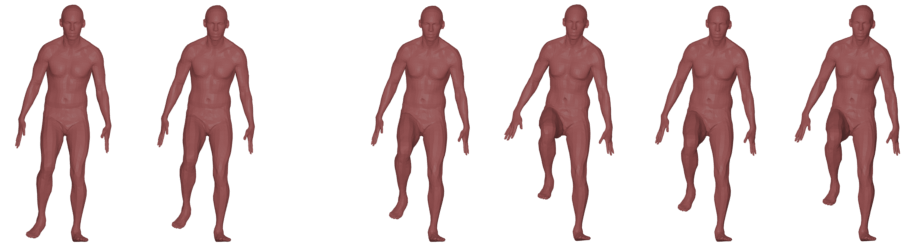}  \\\hline
        \begin{tabular}{l}LIMP \end{tabular}&\includegraphics[width=\linewidth]{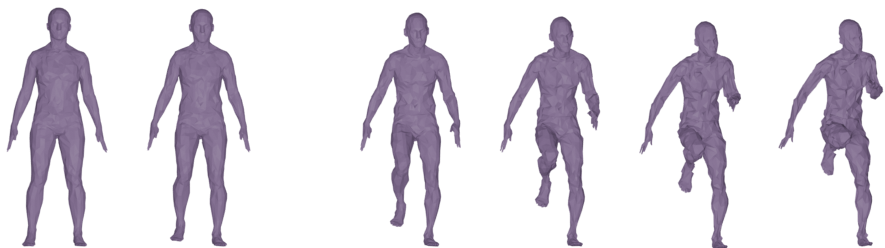}  \\\hline
        \begin{tabular}{l}ARAPreg \end{tabular}&\includegraphics[width=\linewidth]{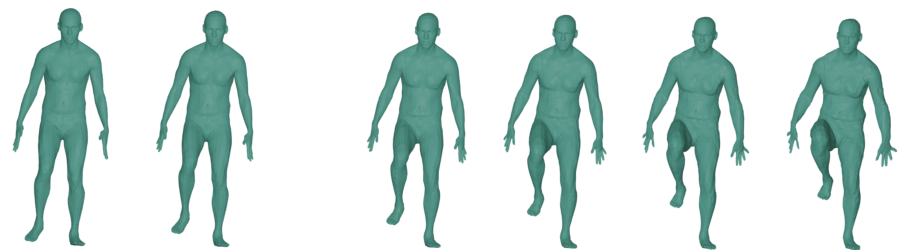} \\\hline
        \begin{tabular}{l}GROUND\\TRUTH\end{tabular}& \includegraphics[width=\linewidth]{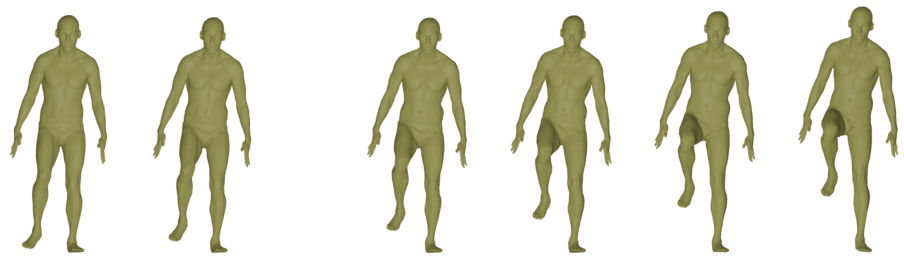}  \\
        &$\underbracket{\hspace*{1.75cm}}_{\text{Input }}\qquad\underbracket{\hspace*{3.5cm}}_{\text{Extrapolation}}$
    \end{tabular}
    \caption{Extrapolation results comparison between our method, LIMP, ARAPReg and D-FAUST Ground Truth. While all methods capture the primary motion of lifting a leg, the extrapolations of LIMP and ARAPreg include extraneous motions of arms and slight changes in body type.}
    \label{fig:extrapolation}
\end{figure}

In the following we consider the shape extrapolation problem, i.e., the task of predicting the future movement given a body shape and an initial movement (deformation). In our Riemannian setting this corresponds to solving the geodesic initial value problem using the method outlined in Section $\ref{sec:model}$. From each of the 10 mini-sequences in our testing set, we recover the latent codes of the first two meshes in the sequence and  then use the first latent code and the difference of the codes as input to our method. In \Cref{fig:extrapolation}, we display the results of our extrapolation method, the extrapolations computed using LIMP and ARAPreg, and the original sequence (see the supplementary material for their corresponding animations). Our method is successful at producing extrapolations that capture the correct motion of the mesh without any extraneous motions that stay in space of human bodies. As with the interpolation comparison, we measure the distance to the ground-truth sequences at each break point and display the results of the quantitative comparison in~\Cref{tab:interp_extrap}. Similar as in the previous experiments, our method significantly outperforms the other methods.

\subsection{Pose and Shape Disentanglement}
\begin{figure}
\centering
\begin{tabular}{m{.8cm}|m{5cm}}
    \includegraphics[width=\linewidth]{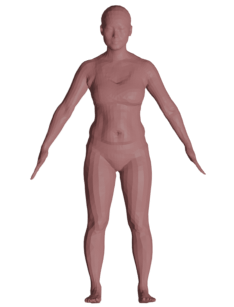}&\includegraphics[width=\linewidth]{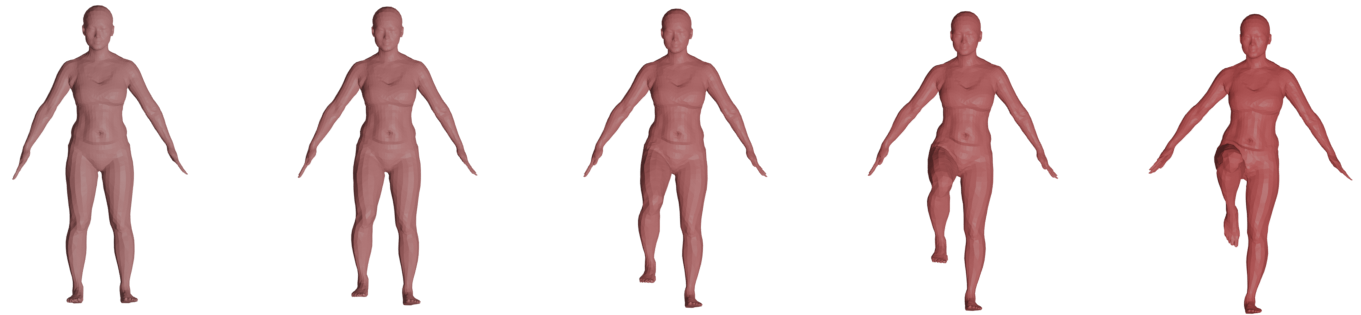}  \\ 
    \includegraphics[width=\linewidth]{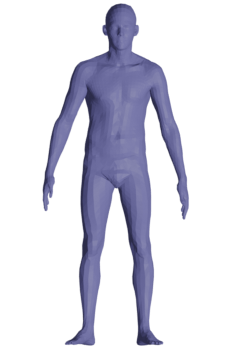}&\includegraphics[width=\linewidth]{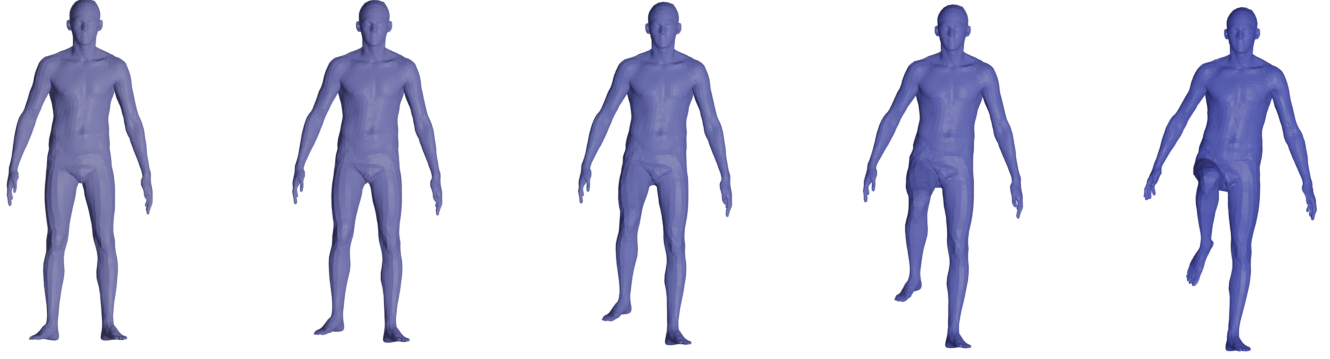}  \\ 
    \includegraphics[width=\linewidth]{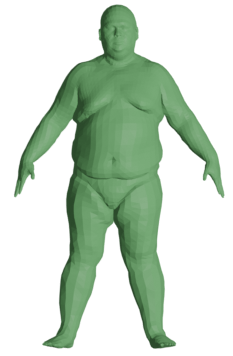}&\includegraphics[width=\linewidth]{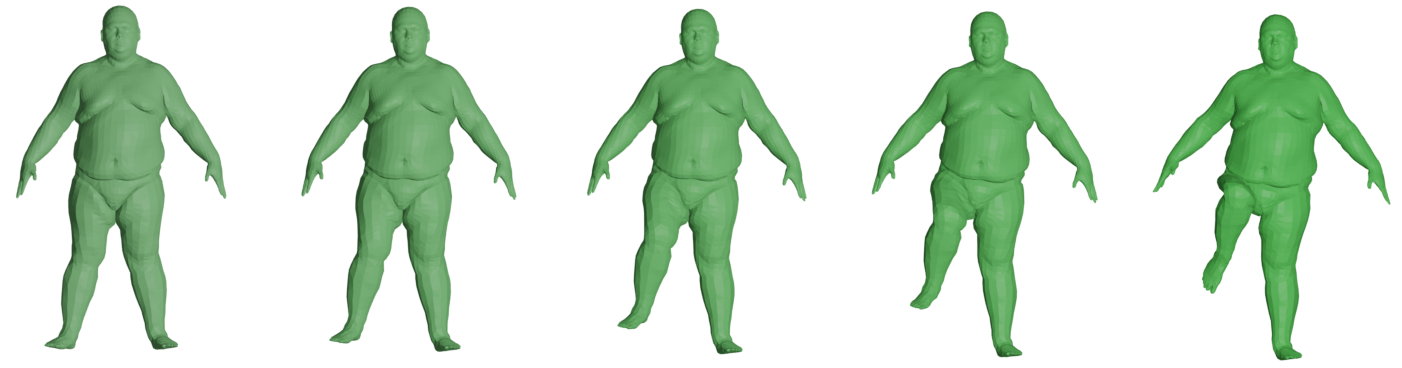}  \\
\end{tabular}
\caption{Motion Transfer: We display the original motion in the top row and the transfer of the motion to the target shapes in the second and third row.}
\label{fig:motion_transfer}
\end{figure}

By the construction of our basis, the first 130 dimensions of our latent space represent the body pose and the remaining 40 dimensions represent the body shape. Thus, our latent codes can be decomposed into coefficients that represent a change in body type and coefficients that represent a change in body pose. As a result, our framework is able to perform instant \textit{\textbf{motion transfer}}: once a motion is represented as a sequence of latent codes we simply replace the shape coefficients of each element of the sequence with the shape coefficients of the target shape. An example of this method in action is displayed in \Cref{fig:motion_transfer}.

\subsection{Random Shape Generation} 
\begin{figure}
\centering
\includegraphics[width=\linewidth]{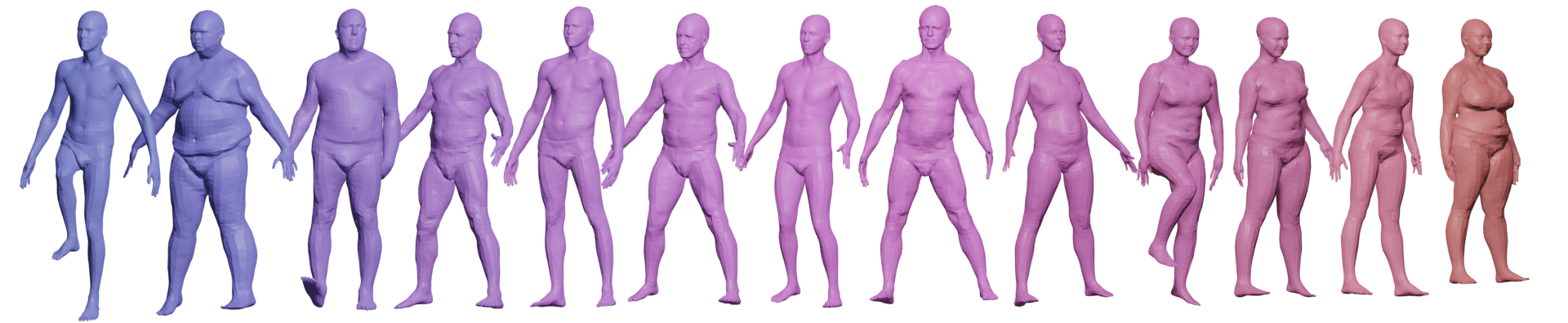}
\caption{13 random shapes generated using a Gaussian mixture model on the space of initial velocities.}\label{fig:random}
\end{figure}

Another possible application of our framework is random shape generation. The idea is to use a data driven distribution on the human shape tangent space. 
Therefore we first perform latent code retrieval on a subset of D-FAUST. We then compute, 
the initial tangent vector of each of these paths in the latent space, separated in pose and shape components.
For each of these collections of tangent vectors, we fit a Gaussian mixture model, which is popular to generate human shapes~\cite{bogo2016keep, omran2018neural}. We used 10 and 6 components respectively, which proved to be sufficient to get visually satisfying random shapes. The generation process consists in sampling a pose and shape vector in the tangent space and to solve the corresponding geodesic initial value problem from the template in the direction of the generated vector. We display a selection of 13 generated shapes in~\Cref{fig:random}.

\section{Conclusions, Limitations and Future work}
In this paper, we proposed a novel framework for 3D human shape interpolation, extrapolation and generation, that relies on learning deformation bases for body type and body pose changes combined with the use of a particular Riemannian structure on the latent space. Importantly, our method  does not require surfaces to have consistent meshes and vertex correspondences and performs well even in the presence of imaging noise such as meshes with holes and/or missing parts. We showcased the different advantages of the proposed framework, in particular over recent deep learning approaches, for recovering and generating meaningful body shape trajectories. We point out, however, that this comes at the price of solving optimization problems to estimate interpolated or extrapolated geodesic paths for our metric. One possible way around this would be to train neural networks in a supervised setting to approximate solutions of those problems, which we leave as a subject of future investigation.  Another potential limitation of the proposed framework is the need for sufficiently rich training data: in the current implementation we used the D-FAUST dataset for the generation of our motion and pose bases. In this dataset there is, however, only very limited movement of the fingers/hands present and consequently our motion basis in not able to faithfully represent such a movement. In future work we plan to use additional datasets for the generation of the bases, which in turn will allow us to get a more complete representation of all possible movements. 

Finally we mention a simple and yet potentially relevant extension of our  model, namely to introduce two distinct Sobolev Riemannian metrics on the shape change and the motion change deformation fields. This comes with the idea of adapting the metric to the different nature of those deformations, and thus even better disentangling these quantities. 

\,\newpage

\appendix
\section*{Appendices}

\section{Formulas and implementations of mesh invariant similarity  metrics}

In the paragraphs below, we add a few details about the similarity metrics used in the registration procedure and for the evaluation and comparison of the different methods. 

First, we remind that the Hausdorff distance between two shapes $[q_0]$ and $[q_1]$ is given by the formula:
\begin{align*}
d_H([q_0], [q_1]) = \max\bigg\{&\sup_{x_0 \in [q_0]} \inf_{x_1\in [q_1]} \|x_0-x_1\|,\\ 
&\sup_{x_1 \in [q_1]}\inf_{x_0 \in [q_0]} \|x_1-x_0\|\bigg\}
\end{align*}
In our numerical experiments, we use the approximate implementation provided by libigl~\cite{libigl}. Note that this metric is typically very sensitive to outliers.

In contrast, the Chamfer distance~\cite{fan2017point, groueix20183d} provides a smoother version of the above and, given two point clouds $[q_0]$ and $[q_1]$, is defined as:
\begin{align*}
d([q_0], [q_1]) = &\frac{1}{N_0}\sum_{x_0 \in [q_0]} \inf_{x_1\in [q_1]} \|x_0-x_1\| \\  
&+ \frac{1}{N_1}\sum_{x_1 \in [q_1]} \inf_{x_0 \in [q_0]} \|x_1-x_0\|.
\end{align*} 
We use the Pytorch implementation of Thibault Groueix\footnote{\url{https://github.com/ThibaultGROUEIX/ChamferDistancePytorch}}. One of the downsides of this metric when applied to discrete surfaces, however, is that it is not necessarily robust to local changes of mesh density (and thus not truly mesh invariant) since it is designed as a distance between point clouds.  

To address that particular issue, as a final measure of reconstruction quality and fidelity metric for the latent code retrieval approach, we rely on the varifold distance introduced in~\cite{charon2013varifold,kaltenmark2017general}. Specifically, assuming the discrete surface $[q_0]$ is given by the reunion of the triangles $\{T_i\}_{i=1,\ldots,F}$ and $[q_1]$ as the reunion of triangles $\{T'_j\}_{j=1,\ldots,F'}$, the discrete approximation of the squared varifold distance writes:
\begin{multline*}
 d^{\text{Var}}([q_0],[q_1])^2 = \sum_{i,j=1}^{F} k(x_i,n_i,x_j,n_j) a_i a_j 
 \\-2\sum_{i,j=1}^{F,F'} k(x_i,n_i,x'_j,n'_j) a_i a'_j
+ \sum_{i,j=1}^{F'} k(x'_i,n'_i,x'_j,n'_j) a'_i a'_j \label{eq:varifold}
\end{multline*}
where $x_i,n_i,a_i$ (resp. $x'_i,n'_i,a'_i$) denote the barycenter, unit normal vector and area of triangle $T_i$ (resp. $T'_i$). Here $k$ is a positive definite kernel function on $\R^3 \times S^2$. While several different families of kernels are possible (see discussion in~\cite{kaltenmark2017general}),  in all the experiments of this paper, we specifically take $k(x,n,x',n') = e^{-\frac{|x-x'|^2}{\sigma^2}}(n\cdot n')^2$ where $\sigma$ can be interpreted as a spatial scale of sensitivity of the metric which is chosen to be quite small ($\sigma=0.025$) in our examples. In this work, we adapted the Python implementation used in $H2\_SurfaceMatch$\footnote{\url{https://github.com/emmanuel-hartman/H2_SurfaceMatch}} which itself relies on the \textit{PyKeops} library \cite{feydy2020fast} for efficient evaluation and automatic differentiation of kernel functions on the GPU.


\section{Second Order Sobolev Metrics}
In Section 3 we outline the desired properties of a metric on the space of immersions that we pullback onto our latent space.  Here we give a more in depth formulation for the family of split second order Sobolev metrics introduced in \cite{hartman2022elastic} that we use in our model.  We begin with a second order metric given by
\begin{equation}\label{eq:second_order}
    \int_\mathcal{T} \langle h,h \rangle +g_q^{-1}(dh,dh) + \langle\Delta_q h,\Delta_q k\rangle\vol_q
\end{equation}
where we view $dh$ as a vector valued one-form, $g_q$ is the pullback of the Euclidean metric on $\R^3$ \textcolor{black}{and $\Delta_q$ denotes the vector Laplace operator induced by the parametrization $q$}. \textcolor{black}{Fixing a coordinate view and treating $d_h$ and $g_q$ as $3\times2$ and $2\times2$ (resp.) matrix fields on $\mathcal{T}$, one can then express $\vol_q$ as $\sqrt{|g_q|}$} and the first order term is computed as $g_q^{-1}(dh,dh)= \operatorname{tr}(dh\cdot g_q^{-1})$. However, using the construction of \cite{Su_2020_CVPRW}, we may further decompose the vector valued one-forms by $dh=  dh_m+dh_+ + dh_\perp+dh_0$. The exact formulas for these terms can be found in \cite{Su_2020_CVPRW}. Each of these components are orthogonal with respect to the metric $g_q^{-1}$. \textcolor{black}{The associated Riemannian energies of the first three can be roughly interpreted from the point of view of linear elasticity \cite{charon2023shape} as measuring the change of metric tensor under constant volume form (shearing), the change in volume density (stretching) and the change of normal vector direction (bending) respectively. The last term doesn't have such a clear interpretation but is necessary to recover the standard first-order Sobolev norm.} Thus we can eventually decompose \Cref{eq:second_order} and introduce nonnegative weighting coefficients, producing the six parameter family of metrics given by
\begin{equation}
\begin{aligned}
&G_q(h,k)=\int_\mathcal{T}\bigg( a_0 \langle h,k \rangle +
a_1 g_q^{-1}(dh_m,dk_m)\\&\qquad\qquad +b_1g_q^{-1}(dh_+,dk_+)+ c_1g_q^{-1}(dh_\bot,dk_\bot)\\&\qquad\qquad+ d_1 g_q^{-1}(dh_0,dk_0)
+a_2 \langle\Delta_q h,\Delta_q k\rangle\bigg)\vol_q.
\end{aligned}
\end{equation}
The weighting of these terms have very natural geometric interpretations which can be adjusted to make the metric more suited for different applications. The zeroth-order term weighted by $a_0$ penalizes how far the surface is moved weighted by the volume form of the surface. The second-order term weighted by $a_2$ penalizes tangent vectors that increase the local curvature of the surface. The interpretation of the first order terms weighted by $a_1, b_1, c_1$ and $d_1$ was discussed above. In the application to human motions, we typically choose $a_1$ and $b_1$ to be the largest coefficients so that our metric penalizes non-isometric motions. In computations, the different terms in the metric are discretized on triangular meshes based on the principles of discrete differential geometry. A review of relevant methods from this field can be found e.g. in \cite{crane2018discrete} and the details of the specific quantities we use are presented in \cite{hartman2022elastic}.

\section{Computational cost}
As stated in the paper, our pipelines are optimization based. We provide a substantial comparison for the different approaches. 
\begin{table}[]
    \centering
    \begin{tabular}{l|c|c|c|c}
    Method & Training & Retrieval & \multicolumn{2}{c}{Interpolation} \\ \hline \hline 
    LIMP & 1.5w & $<$1s & \multicolumn{2}{c}{$<$1s} \\ \hline
    3D-Coded & 12h & 160s & $<$1s & 160s \\\hline
    ARAPReg & 2w & 160s & $<$1s & 160s \\\hline
    BaRe-ESA & $<$1h & 160s & 91s & 160s \\
    \end{tabular}
    \caption{Computation costs for different methods. For the interpolation, the results are as follow: we display on the left the costs in the case latent codes are available, and the cost in the case they're not.}
    \label{tab:cost_train}
\end{table}

All the other approaches require significant training costs compared to BaRe-ESA which requires less than one hour, cf~\Cref{tab:cost_train}. 
On the other hand, BaRe-ESA,  ARAPReg and 3d-Coded require additional optimization  for the latent code retrieval, which we found takes approximately the same time for all three methods. The optimization cost is driven by the mesh invariant costs -- varifold or Chamfer -- which have $n^2$ complexity, where $n$ is the number of vertices. LIMP is the only method that doesn't require optimization, but the network behaves notably bad when the poses are unseen as showed in the experiments. For the interpolation problem our method requires approximately 90 seconds if the latent codes are already available, whereas it takes approximately the same time as one latent code retrieval if they are not available.
All timing results were obtained using a standard home PC with a Intel 3.2 GHz CPU and a GeForce GTX 2070 1620 MHz GPU.

\section{Description of state-of-the-art methods}
We propose a detailed description of the state-of-the-art method we use as baselines. We selected deep learning methods that builds a flat latent space for human shape deformations. They describe as follows:
\begin{itemize}
    \item Learning Latent Shape Representations with Metric Preservation (LIMP) is a deep learning method modeling deformations of shapes using a variational auto encoder with geodesic constraints. The encoder part use a PointNet architecture, which makes it invariant to parameterization. The decoder part is a Multi Layer Perceptron. The geometric constraints are used a loss functions during the training process. The latent space is divided in an extrinsic part and an intrinsic part and the loss is applied on the interpolation in those dimensions. The intrinsic part is constrained using the computation of full geodesic matrix, which make the training process particularly heavy.
    \item As-Rigid-As-Possible Regularization (LIMP) is a deep learning method modeling deformations of shapes using an auto-decoder architecture. The latent codes and the decoder are learned altogether. During the training, an As-Rigid-As-Possible loss is imposed such that the decoder directions are similar to the ARAP ones. This procedure also makes the training procedure heavy. In order to make it parameterization invariant, we replace the $L^2$ metric by the varifold distance, as an alternative to our Riemannian latent space.
    \item 3D correspondences by deep deformation (3D Coded) is a deep learning method modeling deformations of shapes using a variational auto encoder. Similarly to LIMP, the encoder part use a PointNet architecture, which makes it invariant to parameterization. The decoder uses a Multi Layer Perceptron to deform a template mesh, but no constraint is imposed on the interpolation of latent variables. By taking advantage of a high number of training samples ($>200000$), they obtained state-of-the-art results for human shape correspondence.
\end{itemize}
In the paper, all those methods are trained using the same training set as Bare-ESA, from Dynamic FAUST and reported parameters from the respective papers. 
\section{Comparison to the framework of~\cite{hartman2022elastic}}
\begin{figure}
\centering
\includegraphics[width=\linewidth]{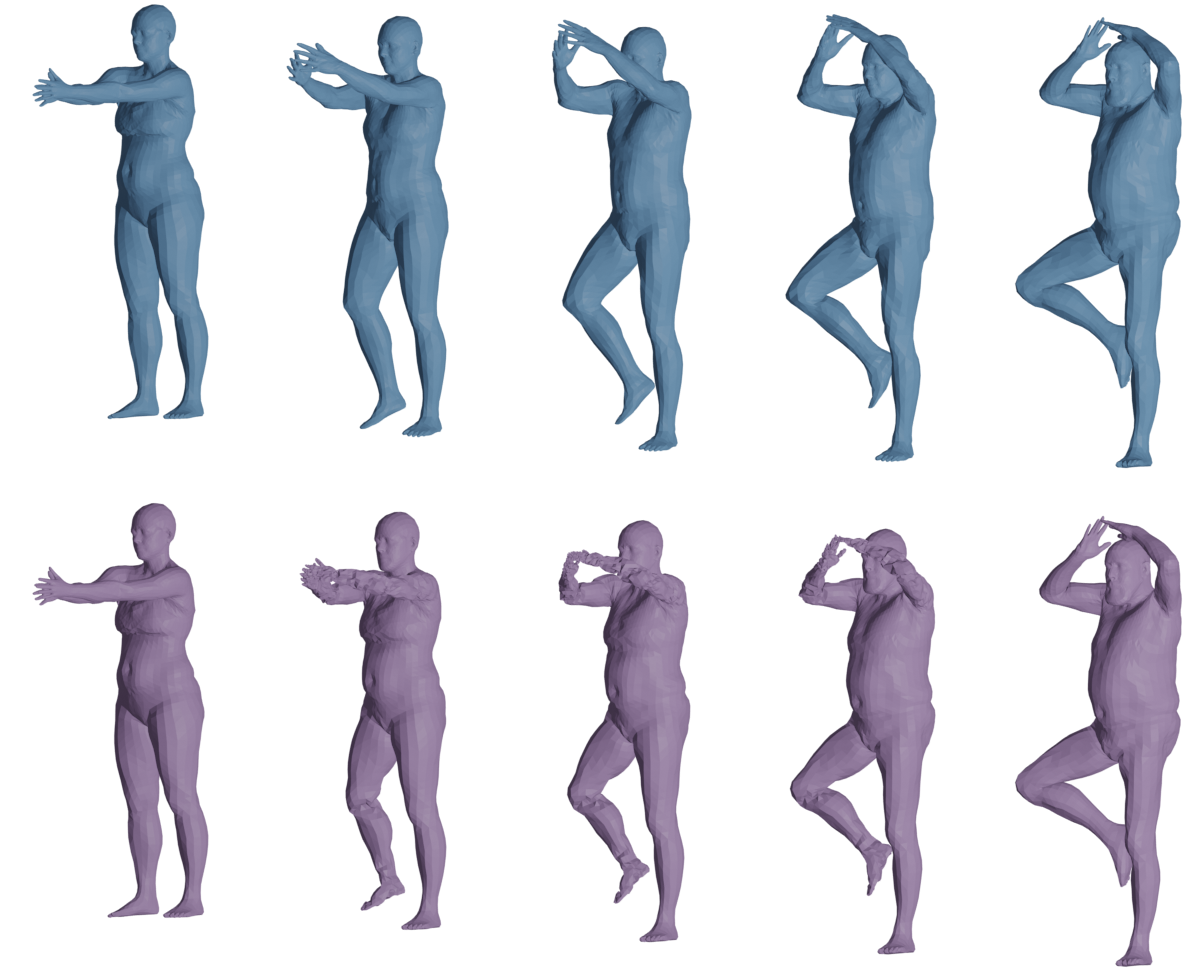}
\caption{First line: optimal deformation calculated using the basis informed ESA of the present article. Second line: optimal deformation calculated using a standard $H^2$-matching.}
\label{fig:basis_vs_nobasis}
\end{figure}

In Figure~\ref{fig:basis_vs_nobasis} we compare BaRe-ESA to the unrestricted method of \cite{hartman2022elastic}. Note, that BaRE-ESA is significantly cheaper to compute as we reduced the dimension of the minimization problem -- the latent space dimension will be in the order of 100s, while the dimension of the unrestricted method is on the order of 10000s. More importantly, one can observe that BaRe-ESA leads to significantly more natural deformations, cf. the movement of the arms in Fig.~\ref{fig:basis_vs_nobasis}.

\section{Algorithmic details}

We provide below the pseudo-code of our algorithm for latent code retrieval of a scan.
\begin{algorithm}
\SetAlgoLined
\KwIn{The target scans $q_0$\;  $a_0, a_1, b_1 c_1, d_1, a_2$ the parameter of the Sobolev elastic metric\; $(\lambda_k, \sigma_k)_{k=0}^p$ the balancing weight and the spatial support of the varifold distance at each refinement step}
\KwOut{$f_{\text{geo}}$: the geodesic connecting $q$ as the coefficients $\alpha$ and the representative $[q_0]$ in the template space.}
 Initialize $\alpha_{ij}=0$ and the path as $\bar q+\sum_{i=1}^m \alpha^j_i h_i+\sum_{i=m+1}^{m+n} \alpha^j_i k_{i-m}.$ \;
 \For{$k\gets0$ \KwTo $p$}{
    Define the energy functional $E(\alpha) = \int_0^1 \overline{G}_{\alpha}(\partial_t\alpha,\partial_t\alpha) dt+\lambda_k \Gamma(F(\alpha)(1),q_0)$ in an automatic differentiation framework (PyTorch here), that computes the gradient value $\nabla_\alpha E$ along the functional value\;
    Minimize $E$ with respect to $\alpha$ with a gradient descent algorithm (SciPy \textit{BFGS} or \textit{L-BFGS-B}), outputing optimal $\alpha_{out}$ coefficients based on initialization $\alpha$\;
    Set $\alpha = \alpha_{out}$\;
 }
 Set $[q_0]$ to be the endpoint of the final geodesic\;
 \Return{$\alpha$ and $[q_0]$}
 \caption{Latent code retrieval of a scan}
 \label{algo:Geodesics}
\end{algorithm}

\section{Comparison against the linear model}

We provide in this section a comparison between the full Bare-ESA model, and the simpler basis restricted linear model (no Riemannian metric). To demonstrate the pertinence of our model, we perform the comparison on the interpolation and extrapolation experiments. We display our results in~\Cref{tab:lin_model}, where we observe a significant gap between linear model and BaRe-ESA respective errors. As expected, the linear model generates non natural deformations, resulting in higher error compared to the real human motions.


\begin{table}[t]
        \centering
        \scriptsize
        \setlength\tabcolsep{2pt} 
        \begin{tabular}{c}
            \begin{adjustbox}{max width=\textwidth}
                \aboverulesep=0ex
                \belowrulesep=0ex
                \renewcommand{\arraystretch}{1.0}
                \begin{tabular}[t]{c|ccc|ccc}
                \multicolumn{1}{c}{} & 
                \multicolumn{3}{c|}{Linear Model} &
                \multicolumn{3}{c}{Bare-ESA} \\
                \cmidrule{2-7}
                & Hausdorff & Chamfer & Varifold  & Hausdorff & Chamfer & Varifold \\
                    \midrule
                    Interpolation &
                    0.18 & 0.07 & 0.03 &
                    \textbf{0.07} & \textbf{0.04} & \textbf{0.02} \\
                     
                    Extrapolation &
                    0.39 & 0.44 & 0.05 & \textbf{0.16} & \textbf{0.10} & \textbf{0.02}\\
            \end{tabular} 
            \end{adjustbox}
          
        \end{tabular}
        \vspace{-3pt}
        \caption{Mean errors of Linear Model and Bare-ESA for interpolation and extrapolation experiments. The Hausdorff, Chamfer and varifold distance are computed against ground truth sequences.}
        \label{tab:lin_model}
        \vspace{-18pt}
\end{table}

\section{Precise requirements of BaRe-ESA}\label{supmat:requirements}
In our experiments we used registered 4D data to construct the body pose basis by performing PCA on the tangent vectors of curves in the space of human body surfaces that represent natural human motions. To construct the body pose basis we compute geodesics between humans in a similar pose and apply the same PCA approach to the tangent vectors of the resulting curves in the space of human body surfaces. We do not consider the use of 4D data for constructing a pose basis as a limitation of our method, but rather an advantage, since it allows us to utilize information from actual human motions to inform the construction of the pose change basis. By contrast, most alternative methods do not utilize the 4D data in this way. Moreover, we can use the framework of~\cite{hartman2022elastic} to construct geodesics as synthetic training data in applications where 4D data is not available. In the same time, the assumption on the existence of humans in a similar body pose could be dropped relatively easily, as one could use the previously obtained body pose basis to create such a training set if needed. This would significantly change the training effort required by our approach.
\end{document}